\documentclass[lettersize,journal]{IEEEtran}
\usepackage{amsmath,amsfonts}
\usepackage{algorithmic}
\usepackage{algorithm}
\usepackage{array}
\usepackage[caption=false,font=normalsize,labelfont=sf,textfont=sf]{subfig}
\usepackage{textcomp}
\usepackage{stfloats}
\usepackage{url}
\usepackage{verbatim}
\usepackage{graphicx}
\usepackage{cite}

\usepackage{amssymb}
\usepackage{hyperref}
\usepackage{multirow}
\usepackage{makecell}
\usepackage{orcidlink}
\usepackage[absolute,overlay]{textpos}
\usepackage[activate={true, nocompatibility}, final, tracking=false, kerning=true, spacing=true, stretch=10, shrink=10]{microtype}

\def\eg{\emph{e.g.}}
\def\ie{\emph{i.e.}}

\def\etal{\emph{et al. }}

\begin{document}

\title{Category-wise Fine-Tuning: \\Resisting Incorrect Pseudo-Labels in Multi-Label Image Classification with Partial Labels}


\author{
Chak Fong Chong \orcidlink{0000-0003-2045-1776},~\IEEEmembership{Student Member,~IEEE},
Xinyi Fang \orcidlink{0009-0000-9823-8612},~\IEEEmembership{Student Member,~IEEE},\\
Jielong Guo \orcidlink{0000-0001-6779-3840},~\IEEEmembership{Senior Member,~IEEE},
Yapeng Wang* \orcidlink{0000-0002-1085-5091},~\IEEEmembership{Member,~IEEE},
Wei Ke \orcidlink{0000-0003-0952-0961},~\IEEEmembership{Member,~IEEE},\\
Chan-Tong Lam \orcidlink{0000-0002-8022-7744},~\IEEEmembership{Senior Member,~IEEE},
Sio-Kei Im \orcidlink{0000-0002-5599-4300},~\IEEEmembership{Member,~IEEE}
\thanks{* Corresponding author (yapengwang@mpu.edu.mo).}
\thanks{Chak Fong Chong, Xinyi Fang, Jielong Guo, Yapeng Wang, Wei Ke, and Chan-Tong Lam are with Faculty of Applied Sciences, Macao Polytechnic University, Macao SAR, China.}
\thanks{Sio-Kei Im is with Macao Polytechnic University, Macao SAR, China.}
}

\markboth{Journal of \LaTeX\ Class Files,~Vol.~14, No.~8, August~2021}%
{Chong \MakeLowercase{\textit{et al.}}: CFT: Resisting Incorrect Pseudo-Labels in Image MLC with Partial Labels}


\maketitle

\begin{abstract}
Large-scale image datasets are often partially labeled, where only a few categories' labels are known for each image.
Assigning pseudo-labels to unknown labels to gain additional training signals has become prevalent for training deep classification models.
However, some pseudo-labels are inevitably incorrect, leading to a notable decline in the model classification performance.
In this paper, we propose a novel method called Category-wise Fine-Tuning (CFT), aiming to reduce model inaccuracies caused by the wrong pseudo-labels.
In particular, CFT employs known labels without pseudo-labels to fine-tune the logistic regressions of trained models individually to calibrate each category's model predictions.
Genetic Algorithm, seldom used for training deep models, is also utilized in CFT to maximize the classification performance directly.
CFT is applied to well-trained models, unlike most existing methods that train models from scratch. Hence, CFT is general and compatible with models trained with different methods and schemes, as demonstrated through extensive experiments.
CFT requires only a few seconds for each category for calibration with consumer-grade GPUs.
We achieve state-of-the-art results on three benchmarking datasets, including the CheXpert chest X-ray competition dataset (ensemble mAUC 93.33\%, single model 91.82\%), partially labeled MS-COCO (average mAP 83.69\%), and Open Image V3 (mAP 85.31\%), outperforming the previous bests by 0.28\%, 2.21\%, 2.50\%, and 0.91\%, respectively.
The single model on CheXpert has been officially evaluated by the competition server, endorsing the correctness of the result.
The outstanding results and generalizability indicate that CFT could be substantial and prevalent for classification model development.
Code is available at: \url{https://github.com/maxium0526/category-wise-fine-tuning}.

\end{abstract}


\section{Introduction}\label{sec:intro}

\begin{textblock*}{20cm}(1.37cm, 0.3cm)
    \scriptsize
    This work has been submitted to the IEEE for possible publication. Copyright may be transferred without notice, after which this version may no longer be accessible.
\end{textblock*}

\begin{figure}
    \centering
    \includegraphics[width=\linewidth]{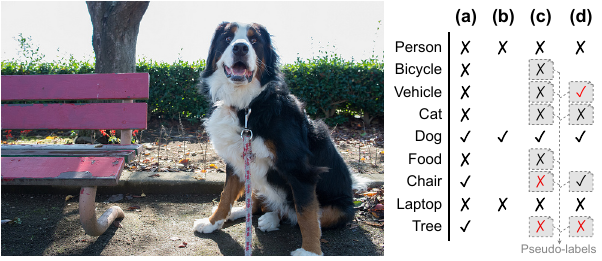}
    \vspace{-18pt}
    \caption{\textbf{(a)} Fully labeled datasets. \textbf{(b)} Partially labeled datasets. Only the labels of some categories are annotated. \textbf{(c)} Assume Negative (AN) assigns negative pseudo-labels for all unknown labels for training. \textbf{(d)} Self-training methods use model predictions as pseudo-labels for unknown labels. Both (c) and (d) inevitably produce incorrect pseudo-labels to the training data, which significantly decrease the classification performance of the trained models.}
    \label{fig:intro}
    \vspace{-12pt}
\end{figure}

\IEEEPARstart{M}{ulti-label} classification (MLC) of images has been a classical computer vision task that recognizes the presence of multiple categories in an image.
It has a wide range of real-world applications, such as medical image interpretation \cite{irvin_chexpert_2019}, self-driving cars \cite{huynh_interactive_2020}, and aerial image analysis \cite{lin_rethinking_2022}.
Recent advances in deep learning have made significant progress in this task, including loss functions \cite{ridnik_asymmetric_2021, kobayashi_two-way_2023}, network architectures \cite{chen_learning_2019, durand_learning_2019}, and prompt learning \cite{sun_dualcoop_2022, hu_dualcoop_2023, ding_exploring_2023}.
The scales of datasets are also rapidly increasing to facilitate the development of high-performance deep classification models, resulting in significant increases in annotation costs.
Therefore, large-scale datasets are often partially labeled, as shown in \autoref{fig:intro}(b). For each image, only the labels of some categories are annotated (known), whereas the rest are unannotated (unknown).

Partially labeled datasets contain fewer labels than fully labeled datasets, resulting in reduced training signals for model training.
Hence, recent methods often assign pseudo-labels to unknown labels, and train the models with the known labels and the pseudo-labels, such as the Assume Negative-based and self-training methods.

Assume Negative (AN) \cite{kim_large_2022, kundu_exploiting_2020, bucak_multi-label_2011, chen_fast_2013, ridnik_asymmetric_2021} assigns negative pseudo-labels to all unknown labels, as shown in \autoref{fig:intro}(c).
It is based on the nature of conventional fully-labeled datasets that negative labels are usually much more than positive labels
(\eg, in the MS-COCO dataset \cite{lin_microsoft_2014}, positive labels account for only 3.6\%, and the rest are all negative labels \cite{ridnik_asymmetric_2021}).
However, it wrongly assigns negative pseudo-labels to the unknown labels that are positive labels (\eg, \autoref{fig:intro} "Chair" and "Tree").
Therefore, some advanced AN methods \cite{ben-baruch_multi-label_2022, kim_large_2022} assign negative pseudo-labels only to those unknown labels likely to be negative labels. Nevertheless, the selections of unknown labels are typically reliant on model predictions, which can be erroneous. Negative pseudo-labels may still be mistakenly assigned to unknown labels that are positive labels.
Hence, AN and its advanced methods inevitably produce incorrect pseudo-labels for model training.

Self-training methods \cite{chen_structured_2022, durand_learning_2019} generate pseudo-labels for unknown labels in a different way from AN.
It first trains a model using the known labels. Then, it uses the model to predict the ground-truths of the unknown labels and uses the predictions as pseudo-labels for the unknown labels. Lastly, the model is re-trained using both known labels and pseudo-labels, as shown in \autoref{fig:intro}(d).
However, model predictions are sometimes incorrect. Using incorrect predictions as pseudo-labels to re-train the model will reinforce the incorrectness, known as confirmation bias \cite{arazo_pseudo-labeling_2020}. The incorrectness is accumulated during training, which can easily form a vicious cycle to generate more incorrect pseudo-labels for the subsequent model training\cite{zhang_enhanced_2016}. 

To summarize, most AN-based (\ie, AN and AN's advanced methods) and self-training methods inevitably produce incorrect pseudo-labels for model training. These incorrect pseudo-labels persistently misguide the model's training, eventually leading to a significant decrease in the model classification performance.

In this paper, we propose a novel method, Category-wise Fine-Tuning (CFT), applied to a model that has been well-trained with an AN-based or self-training method to alleviate the adverse impact of the incorrect pseudo-labels, consequently enhancing the performance of the model.
In particular, CFT only uses the known labels without pseudo-labels to fine-tune the model, enabling the precise calibration of the model to produce more accurate predictions.
Specifically, the model's classification layer is equivalently regarded as a series of logistic regressions (LRs), where each LR outputs the predicted probability of a single category.
Afterwards, each LR is individually fine-tuned using known labels, so the model prediction of every category can be individually calibrated.
Unlike most existing methods that aim to train high-performance models, CFT is applied to well-trained models to improve the performance further, which is rarely seen in the literature on training deep classification models on partially labeled datasets.
In addition, CFT improves the model performance without adding extra network modules to the model. Therefore, it does not increase the model inference cost.

Moreover, we propose two CFT varieties that use two different algorithms to fine-tune each LR: CFT$_\text{BP-ASL}$ uses backpropagation (BP) to minimize the asymmetric loss (ASL) \cite{ridnik_asymmetric_2021}; CFT$_\text{GA}$ uses Genetic Algorithm (GA) \cite{mitchell_introduction_1998}, an evolutionary algorithm rarely used to train deep models, to maximize the performance directly.

We achieve state-of-the-art performances on three diverse benchmarking datasets.
On the \textbf{CheXpert} chest X-ray competition dataset \cite{irvin_chexpert_2019} hosted by Stanford Machine Learning Group, we achieve mAUC 93.33\% (ensemble) and 91.82\% (single model) on the test set, outperforms previous bests by 0.28\% \cite{yuan_large-scale_2021} and 2.21\% \cite{chong_image_2023}, respectively. The single model has been submitted to the competition server for the official evaluation on the test set. The result is available on the competition leaderboard, confirming the correctness of the results.
On the \textbf{MS-COCO} dataset \cite{lin_microsoft_2014} (partially labeled), we achieve average mAP 83.69\%, surpassing the previous best by 2.50\% \cite{yuan_positive_2023}.
On the \textbf{Open Images V3} dataset \cite{kuznetsova_open_2020}, we achieve mAP 85.31\%, surpassing the previous best by 0.91\% \cite{wang_saliency_2023}.

CFT demonstrates its generality and effectiveness across various trained models. Extensive experiments confirm CFT's efficacy for the trained models developed on different datasets (CheXpert, MS-COCO, Open Images V3), different AN-based and self-training methods (AN, LL-R \cite{kim_large_2022}, Selective \cite{ben-baruch_multi-label_2022}, Curriculum Labeling \cite{durand_learning_2019}), and diverse model architectures (convolutional neural networks \cite{he_deep_2016,tan_efficientnetv2_2021, woo_convnext_2023}, Transformers \cite{dosovitskiy_image_2020, liu_swin_2022}). Hence, CFT is promising to become a popular method for developing models on partially labeled datasets.

The main contributions of this paper are summarized:
\begin{enumerate}
    \item We propose Category-wise Fine-Tuning (CFT) that uses known labels to fine-tune each LR of models trained with AN-based methods and self-training methods.
    \item We propose two CFT varieties. CFT$_\text{BP-ASL}$ uses BP to minimize the asymmetric loss \cite{ridnik_asymmetric_2021}. CFT$_\text{GA}$ uses GA \cite{mitchell_introduction_1998}, which is rarely used for training deep models to maximize the performance directly.
    \item We conduct comprehensive experiments and have achieved state-of-the-art performance on three diverse benchmarking datasets, including CheXpert \cite{irvin_chexpert_2019}, MS-COCO \cite{lin_microsoft_2014}, and Open Images V3 \cite{kuznetsova_open_2020}.
\end{enumerate}

This paper is extended from the conference paper entitled \textit{"Category-wise Fine-Tuning for Image Multi-Label Classification with Partial Labels"} \cite{chong_category-wise_2023} with massive additional contents:
\begin{enumerate}
    \item In \cite{chong_category-wise_2023}, CFT is only applied to models trained with AN. This paper extends CFT to models trained with AN-based methods and self-training methods.
    \item In the experiments on MS-COCO in \cite{chong_category-wise_2023}, only ablation studies are conducted. This paper presents a more comprehensive set of experiments on MS-COCO, including ablation studies, evaluations under different model architectures and feature vector's dimensions, and a comparison to state-of-the-art methods.
    \item This paper further conducts experiments on Open Images V3, with an additional analysis of results and a comparison to state-of-the-art methods.
    \item Furthermore, this paper attains state-of-the-art on both MS-COCO and Open Images V3.
    \item Additionally, this paper evaluates the computation time of CFT, a consideration not addressed in \cite{chong_category-wise_2023}.
\end{enumerate}

The rest of this paper is organized as follows.
Sect. \ref{sec:related_work} presents related work on MLC with partial labels.
Sect. \ref{sec:method} describes our proposed methods.
Sect. \ref{sec:experiment} presents the experimental results and discussion.
Sect. \ref{sec:conclusion} concludes this work.

\section{Related Work}\label{sec:related_work}

\subsection{Conventional Approaches}

Cabral \etal \cite{cabral_matrix_2011} formulated an MLC problem under a matrix completion framework and proposed two convex matrix-completion algorithms based on a Rank Minimization criterion.
Xu \etal \cite{xu_speedup_2013} proposed an efficient low-rank matrix completion method with side information for multi-label learning.
Yu \etal \cite{yu_large-scale_2014} proposed a standard error risk minimization framework to learn a linear model with a low-rank constraint.
Yang \etal \cite{yang_improving_2016} additionally used a semantic graph and the semantic graph Laplacian to capture and incorporate the structured semantic correlations between images to help learning.
Additionally, Chen \etal \cite{chen_fast_2013} proposed FastTag, which learns two classification models to predict unknown labels.
Besdies, ML-MG \cite{wu_ml-mg_2015} utilized a mixed graph to jointly incorporate instance-level similarity, class co-occurrence, and semantic label hierarchy to predict unknown labels.
Kapoor \etal \cite{kapoor_multilabel_2012} proposed a Bayesian model for MLC via compressed sensing.

However, the training procedures of \cite{cabral_matrix_2011, xu_speedup_2013, yu_large-scale_2014, yang_improving_2016, chen_fast_2013, kapoor_multilabel_2012} are not well adapted to train deep neural networks. ML-MG \cite{wu_ml-mg_2015} and \cite{yang_improving_2016} require solving convex quadratic optimization problems, which are difficult to scale to large-scale datasets \cite{durand_learning_2019}. Furthermore, most of these methods require putting all training samples into memory for solving optimization problems.
To summarize, these methods are generally not well-suited for training deep models.

\subsection{Deep Learning Approaches}

Binary Relevance \cite{gong_deep_2013, chong_image_2023} casts an MLC problem to multiple individual per-category binary classifications. But each category requires a deep model, causing this approach to be less scalable to large numbers of categories. It also omits the image-image and label-label correlations which may be helpful for model training \cite{durand_learning_2019}.

Several methods use known labels only and ignore all unknown labels to train models. Durand \etal \cite{durand_learning_2019} proposed partial-BCE loss derived from conventional binary crossentropy (BCE) loss. It explicitly ignores the gradient contributions of all unknown labels and normalizes the loss with the known label proportion to give every training sample the same importance.
Similarly, the "No exposure" setting in \cite{kundu_exploiting_2020} ignores the unknown labels in BCE loss.
The P-ASL ($\omega_c=0$) \cite{ben-baruch_multi-label_2022} ignores the unknown labels in the asymmetric loss (ASL) \cite{ridnik_asymmetric_2021}.
Besides, \cite{sun_dualcoop_2022, hu_dualcoop_2023, ding_exploring_2023} adapted large-scale vision-language pre-trained models for MLC with partial labels to leverage its strong alignment of visual and textual feature spaces to compensate the label deficiency.
These methods ensure that the model is trained using clean and accurate labels.
However, some datasets have extremely limited known labels (\eg, in Open Images V3, only 0.086\% of the labels are known) which provide very few training signals. Moreover, it may lead to poor decision boundaries \cite{ben-baruch_multi-label_2022}.

Assume Negative (AN) assigns negative pseudo-labels to all unknown labels.
It is based on the nature of fully labeled datasets that negative labels are usually much more than positive labels.
The "Full exposure" setting in \cite{kundu_exploiting_2020} adapts the BCE loss with this approach to handle unknown labels, and similarly, P-ASL ($\omega_c=1$) in \cite{ben-baruch_multi-label_2022} adapts the asymmetric loss \cite{ridnik_asymmetric_2021}.
AN is simple to implement and provides more labels. However, some unknown labels are positive labels, but AN wrongly assigns negative pseudo-labels to them, harming the model training. Moreover, it aggravates the positive-negative sample imbalance \cite{ridnik_asymmetric_2021}.
Therefore, several methods improve the conventional AN.
Soft exposure \cite{kundu_exploiting_2020} models image and label relationships using knowledge distillation to soften the signals of the negative pseudo-labels.
The Hill loss \cite{zhang_simple_2021}, a re-weighted form of MSE loss, lowers the weight on potentially-incorrect negative pseudo-labels in the shape of a hill to mitigate the impacts.
The Self-Paced Loss Correction (SPLC) method \cite{zhang_simple_2021} adopts a loss derived from the maximum likelihood criterion under an approximate distribution of missing labels to correct pseudo-labels during training gradually.
Kim \etal \cite{kim_large_2022} proposed identifying incorrect negative pseudo-labels through the large loss magnitude. Based on this, they proposed three methods to reject (LL-R), temporarily  correct (LL-Ct), or permanently correct (LL-Cp) the identified pseudo-labels.
BoostLU \cite{kim_bridging_2023} modifies the model to produce class activation maps and element-wisely boosts the attribution scores of the highlighted regions to compensate for the score decrease caused by incorrect negative pseudo-labels.
Selective \cite{ben-baruch_multi-label_2022} assigns negative pseudo-labels to some unknown labels selected based on the conditional probability of the unknown label being positive and the probability of the category being present in an image.
However, these methods usually identify incorrect pseudo-labels based on model predictions during training, which are sometimes incorrect.
This leads to the inclusion of certain unknown labels, specifically positive labels, which are erroneously selected and contribute to the incorporation of negative pseudo-labels.

Self-training methods use model predictions as pseudo-labels for unknown labels.
Durand \etal \cite{durand_learning_2019} proposed a curriculum learning-based method \cite{bengio_curriculum_2009} with a graph neural network that explicitly models the correlations between categories to generate pseudo-labels gradually.
In \cite{huynh_interactive_2020}, the unknown labels of an image are determined by the related labels of semantically similar images. The model is trained interactively with the similarity learner to regularize the predicted labels and learned image features to be smooth on the data manifold.
The adaptive temperature associated model (ATAM) \cite{lin_rethinking_2022} introduces a temperature factor to soften the strict discrepancy between negative and positive labels to classify aerial images.
Structured Semantic Transfer (SST) \cite{chen_structured_2022} and Heterogeneous Semantic Transfer (HST) \cite{chen_heterogeneous_2022} explore the image-specific occurrence and category-specific feature similarities to generate pseudo-labels for unknown labels.
Wang \etal \cite{wang_saliency_2023} proposed a saliency regularization-based framework with class-specific maps to alleviate the dominance of negative labels.
However, self-training methods rely on model predictions, which may occasionally be incorrect, to serve as pseudo-labels for training, which will reinforce the model's tendency to generate consistently inaccurate predictions. This situation is also known as confirmation bias \cite{arazo_pseudo-labeling_2020}. Consequently, this continual training on incorrect pseudo-labels significantly diminishes the model classification performance.

Besides, semantic-aware representation blending (SARB) \cite{pu_semantic-aware_2022} and dual-perspective semantic-aware representation
blending (DSRB) \cite{pu_dual-perspective_2023} perform mixup-like operations across different images in the category-specific representation level to complement unknown labels.
Positive and unlabeled multi-label classification (PU-MLC \cite{yuan_positive_2023}) discards all negative labels and only uses positive and unknown labels for training models.

Unlike most existing methods that aim to train a high-performance model from scratch, our proposed CFT is a method that is applied to well-trained models, which is rarely seen in related work to the best of our knowledge. Moreover, our CFT can further improve the classification performance of models trained using AN-based and self-training methods, all without incurring additional model inference cost. Hence, CFT exhibits promise as a significant and prevalent method in developing models on partially labeled datasets.

\section{Method}\label{sec:method}

\subsection{Problem Formulation}

Considering a multi-label image classification problem with $C$ categories. 
A dataset $\mathcal{D}=\big\{(\mathbf{x}, \mathbf{y})_n\big\}_{n=1}^{N}$ consists of $N$ training samples.
Each sample $(\mathbf{x}, \mathbf{y})$ consists of an image $\mathbf{x}$ and a label vector $\mathbf{y} = [y_1, y_2, y_3, \dots, y_C] \in \{-1, 1, 0\}^{C}$.
The $c^{\text{th}}$ element $y_c$ is the label of category $c$, which is assigned to be either $-1$ (negative), $1$ (positive), or 0 (unknown).

Given a deep classification model $\mathtt{F}_\Theta$ parameterized by $\Theta$ which has already been well-trained on the dataset $\mathcal{D}$ with a AN-based or self-training method. It produces predicted probabilities $\mathbf{p}=\mathtt{F}_\Theta(\mathbf{x})=[p_1, p_2, p_3, \dots, p_C] \in [0, 1]^C$ of an input image $\mathbf{x}$, where the $c^\text{th}$ element $p_c$ is the predicted probability of category $c$.

Since most AN-based and self-training methods inevitably produce incorrect pseudo-labels for model training, as explained in Sect. \ref{sec:intro}. These incorrect pseudo-labels persistently misguide the model training, resulting in a significant decline in the classification performance of the trained model $\mathtt{F}_\Theta$. Our goal is to reduce the negative impact made by the incorrect pseudo-labels to improve the model performance further.

\subsection{Category-wise Fine-Tuning (CFT)} \label{sec:cft}

\begin{figure}
    \centering
    \includegraphics[width=\linewidth]{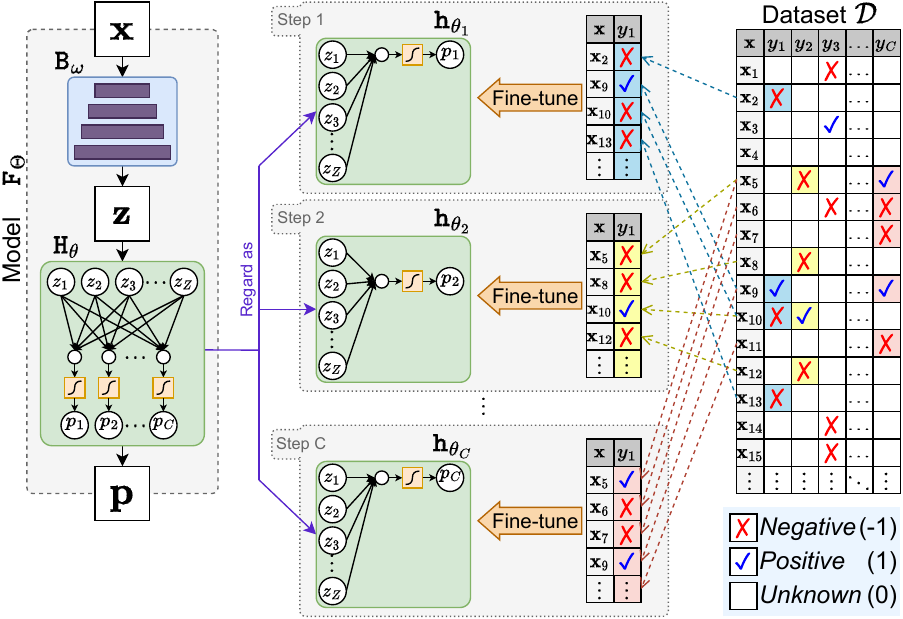}
    \caption{The overview of CFT.}
    \label{fig:cft}
\end{figure}

We propose Category-wise Fine-Tuning (CFT), a method to mitigate the incorrectness of the trained model caused by the incorrect pseudo-labels, as shown in \autoref{fig:cft}.
CFT only uses known labels to fine-tune the model so the model can be precisely calibrated with accurate labels. Thereby, with CFT the model can produce more accurate predictions.

Specifically, the classification layer of the trained model is first equivalently regarded as a series of logistic regressions (LRs) where each LR outputs the predicted probability of one category, as shown in \autoref{fig:cft} (left). Then, each LR is one-by-one fine-tuned using known labels only, as shown in \autoref{fig:cft} (middle). Thus, the model predictions of every category can be individually calibrated.

Given the trained model $\mathtt{F}_\Theta$, we regard its architecture as two components:
\begin{enumerate}
    \item A backbone $\mathtt{B}_\omega$ parameterized by $\omega$ transforms an input image $\mathbf{x}$ to a feature vector $\mathbf{z}=\mathtt{B}_\omega(\mathbf{x}) \in \mathbb{R}^{Z}$. As $\mathtt{B}_\omega$ is always fixed, we simply denote it as $\mathtt{B}$ in the following.
    \item A classification layer $\mathtt{H}_\theta$ which is a fully-connected layer parameterized by $\theta=\{\mathbf{W}, \mathbf{b}\}$ where $\mathbf{W}\in \mathbb{R}^{C\times Z}$ denotes the weights and $\mathbf{b}\in\mathbb{R}^C$ denotes the bias.
$\mathtt{H}_\theta$ transforms a feature vector $\mathbf{z}$ to the predicted probabilities $\mathbf{p}=\mathtt{H}_{\theta}(\mathbf{z}) = \sigma(\mathbf{W}\mathbf{z}^T+\mathbf{b}) \in [0, 1]^C$ where $\sigma(\cdot)$ denotes logistic function activation.
\end{enumerate}

The classification layer $\mathtt{H}_\theta$ is further regarded as a series of $C$ individual logistic regressions (LRs), denoted as $\mathtt{h}_{\theta_1}, \mathtt{h}_{\theta_2}, \mathtt{h}_{\theta_3}, \dots, \mathtt{h}_{\theta_C}$.
The $c^\text{th}$ LR $\mathtt{h}_{\theta_c}$ is parameterized by $\theta_c=\{\mathbf{w}_c, b_c\}$, where $\mathbf{w}_c$ is the $c^\text{th}$ row vector of $\mathbf{W}$ and $b_c$ is the $c^\text{th}$ element of $\mathbf{b}$.
It transforms the feature vector $\mathbf{z}$ to a predicted probability of category $c$ by: $p_c=\mathtt{h}_{\theta_c}(\mathbf{z})=\sigma(\mathbf{w}_c \mathbf{z}^T + b_c)$. Thus, the classification layer can be equivalently represented by the LRs:
\begin{equation}\label{eq:split_to_lr}
    \mathtt{H}_\theta(\mathbf{z})= \big[\mathtt{h}_{\theta_1}(\mathbf{z}), \mathtt{h}_{\theta_2}(\mathbf{z}), \mathtt{h}_{\theta_3}(\mathbf{z}), \cdots, \mathtt{h}_{\theta_C}(\mathbf{z})\big].
\end{equation}

The LRs $\mathtt{h}_{\theta_1}, \mathtt{h}_{\theta_2}, \mathtt{h}_{\theta_3}, \dots, \mathtt{h}_{\theta_C}$ is one-by-one fine-tuned using known labels.
More specifically, the procedure has $C$ steps.
At step $c$ ($c=1,2, \dots, C$), the $c^\text{th}$ LR $\mathtt{h}_{\theta_c}$ is fine-tuned using the training samples in $\mathcal{D}$ that have known labels of category $c$ ($y_c\ne0$).
\ie, the selected samples are $(\mathbf{x}, y_c)\in \{(\mathbf{x}, y_c)|(\mathbf{x}, \mathbf{y})\in \mathcal{D}, y_c\ne0\}$.
The overall procedure of CFT is summarized in Algorithm \autoref{alg:cft}.

\begin{algorithm}
    \caption{CFT}\label{alg:cft}
    \textbf{Input} Trained model $\mathtt{F}_\Theta$, dataset $\mathcal{D}$\\
    \vspace{-14pt}
    \begin{algorithmic}[1]
        \STATE Regard $\mathtt{F}_\Theta$ as a backbone $\mathtt{B}$ and a classification layer $\mathtt{H}_\theta$
        \STATE Regard $\mathtt{H}_\theta$ as LRs $\mathtt{h}_{\theta_1}, \mathtt{h}_{\theta_2}, \mathtt{h}_{\theta_3}, \cdots, \mathtt{h}_{\theta_C}$ (Eq. \ref{eq:split_to_lr})
        \FOR{$c=1, 2,  \dots, C$}
        \STATE Fine-tune $\mathtt{h}_{\theta_c}$ using \scalebox{0.95}[1]{$(\mathbf{x}, y_c)\in \{(\mathbf{x}, y_c)|(\mathbf{x}, \mathbf{y})\in \mathcal{D}, y_c\ne0\}$}
        \ENDFOR
    \end{algorithmic}
    \textbf{Output} Fine-tuned model $\mathtt{F}_\Theta$
\end{algorithm}

CFT is straightforward and generalized. It is applied to a trained model instead of a method used during model training. Hence, It is compatible with models trained with different configurations, such as different AN-based or self-training methods, loss functions, and data augmentation.
Besides, it only fine-tunes the classification layer, the rest of the model (\ie, backbone) is always frozen. Therefore, It is compatible with most model backbones such as CNNs and Transformers.

We propose two CFT varieties that use two different algorithms to fine-tune each LR:
\begin{enumerate}
    \item $\text{CFT}_\text{BP-ASL}$. It uses backpropagation (BP) to minimize each LR's asymmetric loss \cite{ridnik_asymmetric_2021}.
    \item $\text{CFT}_\text{GA}$. It uses Genetic Algorithm (GA) \cite{mitchell_introduction_1998} to maximize each LR's performance.
\end{enumerate}
$\text{CFT}_\text{BP-ASL}$ and $\text{CFT}_\text{GA}$ are described in the following Sect. \ref{sec:cft_bp} and Sect. \ref{sec:cft_ga}, respectively.

\subsection{Minimizing Asymmetric Loss using BP (CFT$_\text{BP-ASL}$)} \label{sec:cft_bp}

In $\text{CFT}_\text{BP-ASL}$, each LR is fine-tuned using backpropagation (BP), like training neural networks. We choose to minimize the asymmetric loss (ASL) \cite{ridnik_asymmetric_2021} to better handle the positive-negative sample imbalance.

For the $c^\text{th}$ LR $\mathtt{h}_{\theta_c}$, it is fine-tuned using BP to minimize
\begin{equation}\label{eq:finetune_lr_with_bp}
    \min_{\theta_c} \sum_{(\mathbf{x}, y_c)\in \{(\mathbf{x}, y_c)|(\mathbf{x}, \mathbf{y})\in \mathcal{D}, y_c\ne0\}} \mathtt{L}_\text{ASL}\Big( y_c, \mathtt{h}_{\theta_c}(\mathtt{B}(\mathbf{x}))\Big)
\end{equation}
\begin{equation} \label{eq:asl}
    \text{where~~}\mathtt{L}_\text{ASL}(y, p)=\begin{cases}
        (1-p)^{\gamma^+}\log(p) & , y=1\\
        (p_m)^{\gamma^-}\log(1-p_m) & , y=-1\\
    \end{cases}
\end{equation}
where $p_m=\max(p-m, 0)$. $\gamma^+, \gamma^-, m$ are hyperparameters.

\subsection{Maximizing Performance using GA (CFT$_\text{GA}$)} \label{sec:cft_ga}

The motivation for proposing $\text{CFT}_\text{GA}$ is the shortcoming of $\text{CFT}_\text{BP-ASL}$ we have encountered in the experiments, as described below.

The performance of a MLC model is usually evaluated by the mean of per-category performance.
For example, the model performance on the CheXpert dataset is evaluated by mAUC (higher is better) which computes the mean of per-category area under the receiver operator characteristic curve (AUC):
$mAUC = \frac{1}{C} \sum_{c=1}^{C} AUC_c$
. $AUC_c$ denotes the AUC of category $c$, which is also the performance of the $c^\text{th}$ LR, as the LR outputs the predicted probabilities of category $c$.
In this case, the goal of CFT is to improve each LR's AUC so that the model performance mAUC can be improved.

In the experiments on the CheXpert dataset, we have applied $\text{CFT}_\text{BP-ASL}$ to a trained model to minimize each LR's ASL (\autoref{eq:finetune_lr_with_bp}). \autoref{fig:chexpert_bp} (left) shows the training's learning curve of $\text{CFT}_\text{BP-ASL}$ fine-tuning the LR of category "Atelectasis".
As can be seen, $\text{CFT}_\text{BP-ASL}$ effectively minimizes the LR's ASL, but the LR's AUC also dramatically decreases as the ASL lowers, meaning $\text{CFT}_\text{BP-ASL}$ worsens the LR's performance instead. A similar phenomenon can also be observed in the validation learning curve shown in \autoref{fig:chexpert_bp} (middle).
After investigation, the cause of this phenomenon is that minimizing crossentropy loss (\eg, BCE loss and ASL) to maximize accuracy \cite{yan_optimizing_2003} does not necessarily lead to the maximizing AUC \cite{yan_optimizing_2003, cortes_auc_2003} or AP \cite{qi_stochastic_2021}, both of which are popular performance metrics for imbalanced data.
Directly maximizing the LR's AUC without minimizing crossentropy loss may avoid this phenomenon. But BP cannot directly maximize AUC since AUC is non-differentiable \cite{yan_optimizing_2003}.

To this end, we propose using Genetic Algorithm (GA) \cite{mitchell_introduction_1998} to fine-tune each LR. We refer to this CFT variety as $\text{CFT}_\text{GA}$.
GA is a global search algorithm inspired by the principle of evolution theory and capable of training neural networks \cite{montana_training_1989,gupta_comparing_1999,david_genetic_2014}.
In nature, individuals better adapted to their environment have higher chances to survive and, consequently, a greater likelihood of producing offspring. This process keeps repeating over generations until the best individual has evolved.
Specifically, GA starts from 1$^\text{st}$ generation with an initial population of individuals, where each individual encodes one possible solution.
In each generation, the individuals are evaluated by a pre-defined fitness function to evaluate how well they fit for the problem.
Individuals with higher fitness have a higher probability of being selected as parents.
Crossover and mutation operations are applied to the parents to produce their offspring to be the next generation's individuals.
The evolution of individuals is repeated over generations until the stop criteria are satisfied (\eg, convergence).

In particular, unlike CFT$_\text{BP-ASL}$ that fine-tunes each LR by minimizing its ASL, CFT$_\text{GA}$ fine-tunes each LR by maximizing its performance, as GA does not require the optimization objective to be differentiable.
Therefore, CFT$_\text{GA}$ can avoid the phenomenon of performance drop caused by minimizing ASL.

For the model trained on the CheXpert dataset, $\text{CFT}_\text{GA}$ fine-tunes each LR to maximize its AUC, referred to as \textbf{$\text{CFT}_\text{GA-AUC}$}.
Similarly, for a model trained on the MS-COCO dataset (model performance metric is mAP). $\text{CFT}_\text{GA}$ maximizes each LR's AP, referred to as \textbf{$\text{CFT}_\text{GA-AP}$}.

In $\text{CFT}_\text{GA}$, the $c^\text{th}$ LR $\mathtt{h}_{\theta_c}$ is fine-tuned using GA to maximize its performance
\begin{equation}\label{eq:finetune_lr_with_ga}
    \max_{\theta_c} \mathtt{M}\bigg( \Big\{\big(y_c, \mathtt{h}_{\theta_c}(\mathtt{B}(\mathbf{x})) \big) \Big| \big(\mathbf{x},\mathbf{y} \big)\in \mathcal{D}, y_c\ne 0 \Big\} \bigg)
\end{equation}
where $\mathtt{M}(\cdot)$ denotes the performance metric which calculates the performance based on the input label-prediction pairs $\{(y, p ) \}$.
For $\text{CFT}_\text{GA-AUC}$, $\mathtt{M}(\cdot)$ is $\mathtt{AUC}(\cdot)$; For $\text{CFT}_\text{GA-AP}$, $\mathtt{M}(\cdot)$ is $\mathtt{AP}(\cdot)$.

We configure GA as follows to optimize \autoref{eq:finetune_lr_with_ga}.
An individual $\mathbf{i}$ represents a possible parameters $\hat{\theta_c}=\{\hat{\mathbf{w_c}}, \hat{b_c}\}$.
For encoding $\mathtt{e}(\cdot)$ from $\hat{\theta_c}$ to $\mathbf{i}$, we set each position in $\mathbf{i}$ to represent one parameter in $\hat{\theta_c}$:
\begin{equation}
    \mathbf{i} = \mathtt{e}(\hat{\theta_c}) = [\hat{w_{c,1}}, \hat{w_{c,2}}, \dots, \hat{w_{c,Z}}, \hat{b_c}] \in \mathbb{R}^{Z+1}
\end{equation}. 
The decoding $\mathtt{d}(\cdot)$ from $\mathbf{i}$ to $\hat{\theta_c}$ is the inverse operation of $\mathtt{e}(\cdot)$.
The fitness of $\mathbf{i}$ is set to be the LR's performance with the decoded parameters $\hat{\theta_c}$:
\begin{equation}
    \mathtt{fit}(\mathbf{i}) = \mathtt{M}\bigg( \Big\{\big(y_c, \mathtt{h}_{\mathtt{d}(\mathbf{i})}(\mathtt{B}(\mathbf{x})) \big) \Big| \big(\mathbf{x},\mathbf{y} \big)\in \mathcal{D}, y_c\ne 0 \Big\} \bigg)\text{.}
\end{equation}

We have compared the effectiveness of $\text{CFT}_\text{BP-ASL}$ and $\text{CFT}_\text{GA}$ in the experiments on three datasets. We also discuss the results in Sect. \ref{sec:discuss_bp_ga}.\\

\vspace{-6pt}
\noindent\textbf{\textit{Remark.}} In the implementations of $\text{CFT}_\text{BP-ASL}$ (\autoref{eq:finetune_lr_with_bp}) and $\text{CFT}_\text{GA}$ (\autoref{eq:finetune_lr_with_ga}), we pre-compute and save $\mathbf{z}=\mathtt{B}(\mathbf{x})$ for every training sample as a "cache" to dramatically reduce the computation time.

\section{Experimental Results and Discussion}\label{sec:experiment}

We conduct comprehensive experiments on three diverse benchmarking datasets, including CheXpert \cite{irvin_chexpert_2019}, MS-COCO (partially labeled versions) \cite{lin_microsoft_2014}, and Open Images V3 \cite{kuznetsova_open_2020}, as summarized in \autoref{tab:dataset}.
The experiments on each dataset is presented in one section. CheXpert is in Sect. \ref{sec:chexpert}, MS-COCO is in Sect. \ref{sec:coco}, and Open Images V3 is in Sect. \ref{sec:oiv3}.

Besides, Sect. \ref{sec:discussion} provides some discussion and Sect. \ref{sec:time_eval} systematically evaluates the computation time of CFT.

\begin{table}[b]
    \caption{The three datasets used in the experiments.}
    \label{tab:dataset}
    \centering
    \scriptsize
    \setlength{\tabcolsep}{3pt}
    \begin{tabular}{cccccc}
        \hline
        Dataset & Image Type & \#Images & \#Categories & Known Label\% & Metric \\
        \hline
        CheXpert & Medical & 224K+ & 14 & 27.17\% & mAUC \\
        MS-COCO & Natural & 120K+ & 80 & 10\%, 20\%, ..., 90\% & mAP \\
        Open Images V3 & Natural & 9M+ & 5,000 & 0.086\% & mAP \\
        \hline
    \end{tabular}
    
\end{table}

\subsection{CheXpert}\label{sec:chexpert}
\subsubsection{Dataset}
CheXpert \cite{irvin_chexpert_2019} is a large-scale chest X-ray image MLC competition dataset introduced by the Stanford Machine Learning Group. It consists of 14 categories of pathologies.
\textbf{The training set} is a partially labeled and consists of 223,414 images. The labels are extracted from the text reports written by radiologists. Besides positive, negative, and unknown labels, the training set also has \textit{uncertain} labels which have different semantic meanings from the unknown labels. An uncertain label captures the uncertainty in diagnosis and ambiguity in the report. An unknown label implies no mentions are found in the report.
Hence, we do not simply treat uncertain labels as unknown labels in the experiments, as described in the following sessions.
\textbf{The validation} and \textbf{test sets} are fully labeled and consist of 234 and 668 images, respectively. The test set is reserved for the competition. Models must be submitted to the competition server for the official evaluation on the test set.

\subsubsection{Performance Metric}

To fairly compare to previous methods, the model performance is evaluated by the official metric, which is the mean of per-category area under the receiver operator characteristic curve (mAUC) of the five selected categories \cite{irvin_chexpert_2019}: Atelectasis, Cardiomegaly, Consolidation, Edema, and Pleural Effusion.

The AUC of a category is computed by \cite{cortes_auc_2003}:
\begin{equation}
    AUC = \frac{1}{{N^+ N^-}} \sum_{i=1}^{N}\sum_{s=1}^{N}\mathbf{I}_{[y^{(i)}=1]}\mathbf{I}_{[y^{(s)}=-1]}\mathbf{I}_{[p^{(i)}>p^{(s)}]}
\end{equation}
given labels $[y^{(1)},\dots, y^{(N)}]\in \{-1, 1\}^N$ and predicted probabilities $[p^{(1)}, \dots, p^{(N)}] \in [0, 1]^N$ of $N$ samples.
$N^+$ and $N^-$ denote the numbers of positive and negative labels, respectively ($N^++N^-=N$).
$\mathbf{I}_{[\cdot]}$ denotes an indicator function.

\subsubsection{Model Training}
We train a model using AN which assigns negative pseudo-labels for all unknown labels, as AN dominates the state-of-the-art results of this dataset.
Uncertain labels are handled by U-Ones+LSR method \cite{pham_interpreting_2021}. \ie, each uncertain label is treated as a random scalar between $[0.55, 0.85]$.
DenseNet-121 \cite{huang_densely_2017} pre-trained on ImageNet-1K \cite{deng_imagenet_2009} is used as the model backbone.
Training images are augmented by horizontal flip, rotate $\pm20^{\circ}$, and scale $\pm3\%$, then resized to $224^2$ and rescaled to $[0, 1]$ \cite{chong_gan-based_2021,chong_image_2023}. BCE loss, batch size 32 and Adam ($lr=1\times10^{-4}$) \cite{kingma_adam_2014} are used to update parameters for 10 epochs.

The trained model achieves mAUC 89.62\% on the validation set, which already outperforms the single model of $\text{2}^{\text{nd}}$ place (89.40\%) on the leaderboard \cite{pham_interpreting_2021}, meaning the trained model is not underfitting.

\subsubsection{Ablation Studies}

To study the effectiveness of the proposed CFT, we respectively apply $\text{CFT}_\text{BP-ASL}$ and $\text{CFT}_\text{GA-AUC}$ to the trained model.

In the fine-tuning of each LR, uncertain labels are treated as unknown labels.
For $\text{CFT}_\text{BP-ASL}$, we set ASL's $\gamma^+=\gamma^-=m=0$ (equivalent to BCE loss). Full-batch gradient descent with learning rate $1\times10^{-4}$ is used to update parameters for 500 epochs.
For $\text{CFT}_\text{GA-AUC}$, we use the GA implementation of PyGAD \cite{gad_pygad_2021}.
30 individuals initialized by encoding the original parameters are evolved over 500 generations.
Roulette wheel selection is used to select 14 individuals as parents.
10 of the parents are additionally kept as individuals in the next generation.
2-point crossover is used with a probability of 80\%.
Mutation probability is set to be 2\%.
When a mutation occurs, $1\%$ of the positions are mutated by adding random scalars drawn from $[-0.02, 0.02]$.

\autoref{tab:chexpert_ablation_studies} reports the results on the validation set.
Both $\text{CFT}_\text{BP-ASL}$ and $\text{CFT}_\text{GA-AUC}$ improve the AUCs of all 5 categories of the trained model.
$\text{CFT}_\text{GA-AUC}$ achieves mAUC +1.89\% which is much larger than $\text{CFT}_\text{BP-ASL}$ (+0.27\%).

\begin{figure}
    \centering
    \includegraphics[width=\linewidth]{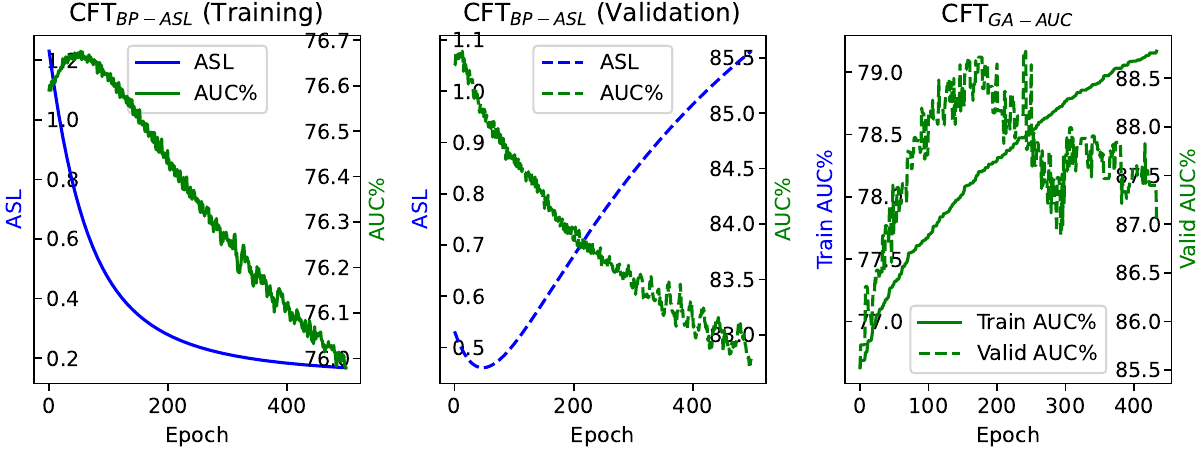}
    \caption{Learning curves of $\text{CFT}_\text{BP-ASL}$ (\textbf{left} and \textbf{middle}) and $\text{CFT}_\text{GA-AUC}$ (\textbf{right}) fine-tuning the LR of the category "Atelectasis".}
    \label{fig:chexpert_bp}
\end{figure}

Particularly, on the category "Atelectasis", $\text{CFT}_\text{BP-ASL}$ only achieves improvement of AUC +0.04\%, which is much less than $\text{CFT}_\text{GA-AUC}$ (+3.26\%).
To analyze this result, we plot the learning curves of $\text{CFT}_\text{BP-ASL}$ and $\text{CFT}_\text{GA-AUC}$ on the category's LR.
\autoref{fig:chexpert_bp} (left and middle) show the training and validation curves of $\text{CFT}_\text{BP-ASL}$, respectively. In training, $\text{CFT}_\text{BP-ASL}$ effectively decreases ASL, but also leads to the decrease of AUC.
Similarly, in validation, AUC almost always decreases even if ASL decreases.
On the other hand, \autoref{fig:chexpert_bp} (right) shows the curves of $\text{CFT}_\text{GA-AUC}$. In comparison to $\text{CFT}_\text{BP-ASL}$, $\text{CFT}_\text{GA-AUC}$ can effectively increase the training AUC. It is because $\text{CFT}_\text{GA-AUC}$ can directly maximize AUC, which avoids minimizing ASL causing AUC to decrease. Moreover, the increase in training AUC also leads to increased validation AUC.\\

\begin{table}[]
    \caption{Ablation studies on CheXpert validation set. In AUC\%($\uparrow$). (\textbf{Best} is in bold, \underline{second-best} is underlined.)}
    \label{tab:chexpert_ablation_studies}
    \centering
    \begin{tabular}{lcccccc}
            \hline
            Method & Ate. & Car. & Con. & Ede. & P.E. & \textit{Mean} \\
            \hline
            AN & 85.52 & 84.18 & 93.34 & 92.73 & 92.34 & \textit{89.62} \\
            \quad+$\text{FT}_\text{BP-ASL}$ & 85.74 & 84.04 & 93.33 & 92.81 & 92.38 & \textit{89.66} \\
            \quad+$\text{CFT}_\text{BP-ASL}$ & 85.56 & 84.99 & 93.53 & 92.87 & 92.47 & \textit{89.89} \\
            \quad+$\text{CFT}_\text{BP-WMW}$ & 87.24 & \underline{87.92} & \textbf{94.69} & 92.87 & \underline{92.49} & \textit{91.04} \\
            \quad+$\text{CFT}_\text{BP-AUCM}$ & \textbf{89.13} & 87.67 & 93.77 & \textbf{93.21} & 92.40 & \underline{\textit{91.23}} \\
            \quad+$\text{CFT}_\text{GA-AUC}$ & \underline{88.78} & \textbf{88.56} & \underline{94.53} & \underline{93.01} & \textbf{92.69} & \textit{\textbf{91.51}} \\
            \hline
        \end{tabular}
\end{table}

\noindent\textbf{Evaluation of One-by-One Fine-Tuning.}
The key characteristic of CFT is the one-by-one fine-tuning of the LRs.
To analyze the its effectiveness, we compare $\text{CFT}_\text{BP-ASL}$ to fine-tuning the whole classification layer using BP and ASL (referred to as $\text{FT}_\text{BP-ASL}$).

$\text{FT}_\text{BP-ASL}$ only improves mAUC by +0.04\%, much less than $\text{CFT}_\text{BP-ASL}$ (+0.27\%). The AUCs of Cardiomegaly and Consolidation are even dropped.
Because in $\text{FT}_\text{BP-ASL}$, the 5 LRs achieve their highest validation AUCs at different epochs (at 628, 132, 896, 79, 376, respectively).
However, $\text{FT}_\text{BP-ASL}$ early-stops at the epoch with highest validation mAUC (epoch 608).
Therefore, $\text{FT}_\text{BP-ASL}$ cannot take the highest validation AUCs for all the 5 LRs.
On the other hand, $\text{CFT}_\text{BP-ASL}$ one-by-one fine-tunes the 5 LRs. The fine-tuning of every LR can early-stop at its best epoch to take their highest validation AUCs.
Therefore, $\text{CFT}_\text{BP-ASL}$ is much effective than $\text{FT}_\text{BP-ASL}$, demonstrating the effectiveness of the one-by-one fine-tuning.\\

\noindent\textbf{Alternative Losses.}
In the literature, we found several loss functions \cite{yan_optimizing_2003, yuan_large-scale_2021} designed to particularly maximize AUC, which may be ideal alternatives to ASL to fine-tune LRs.
Hence, we additionally compare two more CFT varieties that fine-tune each LR by minimizing two different AUC losses using BP:
\begin{enumerate}
    \item $\text{CFT}_\text{BP-WMW}$. It uses BP to minimize the loss proposed in \cite{yan_optimizing_2003}. Stochastic gradient descent, learning rate $1\times 10^{-3}$, momentum $0.9$, and batch size 32768 are used. 
    \item $\text{CFT}_\text{BP-AUCM}$. It uses BP to minimize the AUC margin loss \cite{yuan_large-scale_2021}. We follow the original paper \cite{yuan_large-scale_2021} to use PESG \cite{guo_fast_2020}. Learning rate $1\times10^{-2}$, margin $1$, and full batch are set.
\end{enumerate}
The results are also shown in \autoref{tab:chexpert_ablation_studies}. $\text{CFT}_\text{BP-WMW}$ and $\text{CFT}_\text{BP-AUCM}$ improve the mAUC by +1.42\% and +1.61\%, respectively, which are much more effective than $\text{CFT}_\text{BP-ASL}$ (+0.27\%).
Surprisingly, $\text{CFT}_\text{GA-AUC}$ (+1.89\%) is more effective than $\text{CFT}_\text{BP-WMW}$ and $\text{CFT}_\text{BP-AUCM}$, probably because GA may demonstrate a higher capability in identifying near global optima compared to BP.

\subsubsection{Exploiting Uncertainty Labels}
In the ablation studies mentioned above, the uncertain labels are treated as unknown labels. However, it may be sub-optimal, as reported in the previous works on this dataset \cite{pham_interpreting_2021, irvin_chexpert_2019}.
Therefore, we compare three methods to handle uncertain labels in $\text{CFT}_\text{GA-AUC}$: treat as unknown labels (same as in ablation studies), positive labels \cite{irvin_chexpert_2019}, and negative labels \cite{irvin_chexpert_2019}, referred to as U-Ignore, U-Ones, and U-Zeros, respectively.

The result is shown in \autoref{tab:chexpert_uncertain}. The three methods achieve different AUCs on the 5 categories. Fortunately, in CFT, each LR is individually fine-tuned.
Hence, when fine-tuning each LR, we can use its best methods to handle uncertain labels.
For example, when fine-tuning the LR of Atelectasis, we use U-Ignore, as it achieves the highest AUC. Similarly, we choose U-Ones for the LR of Edema.
We refer to the fine-tuned model developed in this "greedy" way as AN+CFT$_\text{GA-AUC}$(Greedy). It achieves mAUC 91.78\%, which is +2.16\% higher than the trained model before applying CFT (89.62\%).

\begin{table}[]
    \caption{Exploiting Uncertainty Labels. Results on CheXpert validation set. In AUC\%($\uparrow$).}
    \label{tab:chexpert_uncertain}
    \centering
    \setlength{\tabcolsep}{4pt}
    \begin{tabular}{lcccccc}
            \hline
            Method & Ate. & Car. & Con. & Ede. & P.E. & \textit{Mean}\\
            \hline
            AN & 85.52 & 84.18 & 93.34 & 92.73 & 92.34 & \textit{89.62} \\
            \quad+$\text{CFT}_\text{GA-AUC}$(U-Ignore) & \textbf{88.78} & \textbf{88.56} & 94.53 & 93.01 & 92.69 & \textit{91.51} \\
            \quad+$\text{CFT}_\text{GA-AUC}$(U-Ones) & 88.56 & 87.91 & 93.83 & \textbf{93.10} & \textbf{92.84} & \textit{91.25} \\
            \quad+$\text{CFT}_\text{GA-AUC}$(U-Zeros) & 85.47 & 88.19 & \textbf{95.62} & 93.00 & 92.41 & \textit{90.94} \\
            \quad+\textbf{$\text{CFT}_\text{GA-AUC}$(Greedy)} & \textbf{88.78} & \textbf{88.56} & \textbf{95.62} & \textbf{93.10} & \textbf{92.84} & \textbf{\textit{91.78}} \\
            \hline
        \end{tabular}
    
\end{table}

\subsubsection{Comparison to State-of-the-art}
\label{sec:chexpert_test}
We compare AN+CFT$_\text{GA-AUC}$(Greedy) to the state-of-the-art methods on the \textbf{test} set. Most methods treat unknown labels as negative labels and hence can be considered as strong baselines of AN. \autoref{tab:chexpert_sota} shows the comparisons of single models and ensembles.

\begin{table}[]
    \caption{Comparison to state-of-the-art on CheXpert test set. In AUC\%($\uparrow$).}
    \label{tab:chexpert_sota}
    \setlength{\tabcolsep}{3pt}
    \begin{tabular}{lcccccc}
        \hline
        \multicolumn{7}{c}{\textit{\textbf{Single Model}}}\\
        \hline
        Rank \& Method & Ate. & Car. & Con. & Ede. & P.E. & \textit{Mean} \\
        \hline
        147 Chong \etal \cite{chong_gan-based_2021} & 85.67 & 89.30 & 82.15 & 90.92 & 95.56 & \textit{88.72} \\
        151 Multiview (R50) \cite{jansson_multi-view_2021} & 85.60 & 90.85 & 81.07 & 89.45 & 95.85 & \textit{88.60} \\
        134 Multiview (D121) \cite{jansson_multi-view_2021} & 86.49 & \textbf{90.95} & 83.99 & 89.62 & \textbf{96.34} & \textit{89.50} \\
        127 PTRN \cite{chong_image_2023}  & 85.66 & 89.06 & 86.89 & 90.94 & 95.47 & \textit{89.61} \\
        \phantom{0}83 Anatomy-XNet* \cite{kamal_anatomy_2021} & 89.08 & 89.78 & 89.57 & 92.15 & 95.17 & \textit{91.15} \\
        \phantom{0}53 \scalebox{0.77}[1]{\textbf{(Ours) AN+CFT$_\text{GA-AUC}$(Greedy)}} & \textbf{88.58} & 90.20 & \textbf{90.99} & \textbf{93.06} & 96.26 & \textit{\textbf{91.82}}\\
        \hline
        \hline
        \multicolumn{7}{c}{\textit{\textbf{Ensemble}}}\\
        \hline
        Rank \& Method & Ate. & Car. & Con. & Ede. & P.E. & \textit{Mean} \\
        \hline
        102 PTRN \cite{chong_image_2023} & 85.73 & 89.90 & 90.57 & 91.66 & 95.04 & \textit{90.58} \\
        \phantom{0}98 Stanford Baseline \cite{irvin_chexpert_2019} & 85.50 & 89.77 & 89.76 & 91.56 & 96.67 & \textit{90.65} \\
        \phantom{0}14 Anatomy-XNet* \cite{kamal_anatomy_2021} & 89.16 & 93.79 & 91.02 & 93.27 & 95.97 & \textit{92.64} \\
        \phantom{0}\phantom{0}5 YWW \cite{ye_weakly_2020} & 88.18 & \textbf{93.96} & \textbf{93.43} & 92.72 & 96.15 & \textit{92.89} \\
        \phantom{0}\phantom{0}4 HL-v4 \cite{pham_interpreting_2021} & 89.42 & 93.63 & 92.19 & 92.70 & 96.65 & 92.92 \\
        \phantom{0}\phantom{0}3 CT-LSR & 90.18 & 92.83 & 91.84 & 93.13 & 96.65 & 92.92 \\
        \phantom{0}\phantom{0}2 HL-v1 \cite{pham_interpreting_2021} & 90.13 & 93.18 & 92.11 & 92.89 & \textbf{96.68} & \textit{93.00} \\
        \phantom{0}\phantom{0}1 DAM \cite{yuan_large-scale_2021} & 88.65 & 93.72 & 93.21 & 93.00 & 96.64 & \textit{93.05} \\
        \phantom{0}\phantom{0}- \scalebox{0.77}[1]{\textbf{(Ours) AN+CFT$_\text{GA-AUC}$(Greedy)}} & \textbf{91.52} & 93.73 & 91.57 & \textbf{93.33} & 96.50 & \textit{\textbf{93.33}} \\
        \hline
    \end{tabular}
    \vspace{-2pt}
    
    \footnotesize{~* Use extra training data.}
\end{table}

\noindent\textbf{Single Model.} We have submitted AN+CFT$_\text{GA-AUC}$(Greedy) to the competition server for the official evaluation on the test set. The result is available in the leaderboard\footnote{Leaderboard: \url{https://stanfordmlgroup.github.io/competitions/chexpert/}.} and ranks \#53. AN+CFT$_\text{GA-AUC}$(Greedy) achieves mAUC 91.82\%, which is the highest single model AUC in the leaderboard and literature, to our knowledge. Moreover, it outperforms Anatomy-XNet (91.15\%) \cite{kamal_anatomy_2021}, the second-best single model in the leaderboard that even used extra segmentation masks in training.

\noindent\textbf{Ensemble.}
We build an ensemble composed of AN+CFT$_\text{GA-AUC}$(Greedy) and another 4 single models. Similar to $2^\text{nd}$ on the leaderboard \cite{pham_interpreting_2021}, we use test time augmentation \cite{simonyan_very_2014} for more robust predictions: scale $\pm 5\%$, rotate $\pm 5^{\circ}$, and translate $\pm 5\%$.
Since the competition was suddenly closed, our ensemble cannot be submitted to the competition server. After the test set is released and can be downloaded, we evaluate our ensemble on a local machine. 
As shown in \autoref{tab:chexpert_sota}, the top 5 spots on the leaderboard achieved very close mAUC (92.89\% to 93.05\%). Therefore, we consider 0.2\% improvement to be significant. Our ensemble achieves mAUC 93.33\%, which significantly surpasses the previous best (DAM \cite{yuan_large-scale_2021}) by 0.28\% and others methods.

\begin{table}
    \caption{The overfitting on the validation set. In mAUC\%.}
    \label{tab:chexpert_overfitting}
    \centering
    \setlength{\tabcolsep}{5pt}
    \begin{tabular}{lccc}
        \hline
        Method & Valid. & Test & $\Delta$ \\
        \hline
        Chong et al. \cite{chong_gan-based_2021} & 88.9 & 88.7 & \textcolor{red}{-0.2} \\
        Multiview(R50) \cite{jansson_multi-view_2021} & 90.4 & 88.6 & \textcolor{red}{-1.8} \\
        Multiview(D121) \cite{jansson_multi-view_2021} & 90.1 & 89.5 & \textcolor{red}{-0.6} \\
        HL-v1 \cite{pham_interpreting_2021} & 94.0 & 93.0 & \textcolor{red}{-1.0} \\
        Stanford Baseline \cite{irvin_chexpert_2019} & 90.6 & 90.7 & \textcolor{green}{+0.1} \\
        \textbf{(Ours) AN+CFT$_\text{GA-AUC}$(Greedy)} & \textbf{91.8} & \textbf{91.8} & \textbf{\textcolor{green}{+0.0}} \\
        \hline
    \end{tabular}
\end{table}

We also assess whether CFT would cause overfitting on the validation set (\ie, validation mAUC is higher than test mAUC). To this end, we compare the overfitting of AN+CFT$_\text{GA-AUC}$(Greedy) to other methods. As shown in \autoref{tab:chexpert_overfitting}, overfitting on the validation set is not observed in CFT.

\subsection{MS-COCO (partially labeled versions)}\label{sec:coco}

\subsubsection{Dataset}
MS-COCO \cite{lin_microsoft_2014} (2014 split) is a large-scale computer vision dataset. Under the MLC settings, it is a fully labeled MLC dataset with 80 categories. The training and validation sets have around 80k and 40k images, respectively.

As the training set is fully labeled, we randomly drop some labels to simulate partially labeled datasets. To facilitate comparisons with previous methods, we randomly drop 90\%, 80\%, ..., and 10\% of the labels to simulate known label proportions of 10\%, 20\%, ..., and 90\%, respectively.

\subsubsection{Performance Metric}

The model performance is evaluated by the mean of per-category AP (mAP). The AP of a category is computed by \cite{qi_stochastic_2021}:
\begin{equation}\label{eqn:ap}
    AP = \frac{1}{N^+} \sum_{i=1}^{N} \mathbf{I}_{[y^{(i)}=1]}\frac{\sum_{s=1}^{N} \mathbf{I}_{[y^{(s)}=1]}\mathbf{I}_{[p^{(i)} \geq p^{(s)}]}}{\sum_{s=1}^{N} \mathbf{I}_{[p^{(i)} \geq p^{(s)}]}}
\end{equation}
. We focus on average mAP under known label proportions of 10\%, 20\%, ..., and 90\%.

\subsubsection{Model Training} \label{sec:mscoco_training}
To demonstrate CFT on models trained with different methods generating pseudo-labels, we train 5 models using 5 different methods, respectively:
\begin{enumerate}
    \item \textbf{Assume Negative (AN)}. It assigns negative pseudo-labels to all unknown labels.
    \item \textbf{Large Loss Rejection (LL-R)} \cite{kim_large_2022} is derived from AN. It removes the negative pseudo-labels of the unknown labels with large losses before computing gradients.
    \item \textbf{Selective} \cite{ben-baruch_multi-label_2022} is derived from AN. It assigns negative pseudo-labels to some unknown labels selected based on the label prior and the label likelihood.
    \item \textbf{Curriculum Labeling with score threshold strategy (CL$_\text{score}$)} \cite{durand_learning_2019} is a self-training method that gradually generate easy positive and negative pseudo-labels based on classification score.
    \item \textbf{Curriculum Labeling with predicting only positive labels (CL$_\text{score+}$)} \cite{durand_learning_2019} is similar to CL$_\text{score}$ but only positive pseudo-labels are generated.
    
\end{enumerate}

\begin{table*}[]
    \caption{Ablation studies on MS-COCO. In mAP\%($\uparrow$).}
    \label{tab:mscoco_ablation}
    \centering
    \begin{tabular}{lcccccccccc@{\hspace{-0.6\tabcolsep}}c}
        \hline
        \multirow{2}{*}{Method} & \multicolumn{9}{c}{Known Label Proportion} & \multirow{2}{*}{\textit{Average}} & \\
        \cline{2-10}
         & 10\% & 20\% & 30\% & 40\% & 50\% & 60\% & 70\% & 80\% & 90\% & \\
         \hline
         
         AN & 66.77 & 72.10 & 75.29 & 77.18 & 78.45 & 79.90 & 80.98 & 82.48 & 83.77 & \textit{77.44} \\
         \quad+$\text{CFT}_\text{BP-ASL}$ & \textbf{70.83} & \textbf{74.61} & \textbf{77.40} & \textbf{78.65} & \textbf{79.82} & \textbf{80.73} & \textbf{81.44} & \textbf{82.61} & 83.83 & \textit{\textbf{78.88}} & $^{+1.44}$ \\
         \quad+$\text{CFT}_\text{GA-AP}$ & 67.26 & 72.49 & 75.81 & 77.60 & 78.86 & 80.26 & 81.21 & 82.57 & \textbf{83.85} & \textit{77.77} & $^{+0.33}$\\
         \hline

         LL-R \cite{kim_large_2022} & 68.35 & 75.91 & 79.33 & 81.15 & 82.13 & 82.89 & 83.39 & 83.84 & 84.15 & \textit{80.13} \\
        \quad+$\text{CFT}_\text{BP-ASL}$ & \textbf{72.18} & \textbf{77.46} & \textbf{80.39} & \textbf{81.94} & \textbf{82.78} & \textbf{83.52} & \textbf{83.93} & \textbf{84.53} & \textbf{84.85} & \textit{\textbf{81.29}} & $^{+1.16}$\\
        \quad+$\text{CFT}_\text{GA-AP}$ & 70.55 & 76.95 & 80.01 & 81.69 & 82.41 & 83.18 & 83.63 & 84.16 & 84.43 & \textit{80.78} & $^{+0.65}$\\

        \hline

        Selective \cite{ben-baruch_multi-label_2022} & 77.69 & 80.62 & 82.08 & 83.04 & 83.70 & 84.32 & 84.67 & 84.99 & 85.36 & \textit{82.94}  \\
        \quad+$\text{CFT}_\text{BP-ASL}$ & \textbf{78.02} & \textbf{80.91} & \textbf{82.35} & \textbf{83.30} & \textbf{83.90} & \textbf{84.50} & \textbf{84.84} & \textbf{85.12} & \textbf{85.49} & \textit{\textbf{83.16}} & $^{+0.22}$\\
        \quad+$\text{CFT}_\text{GA-AP}$ & 77.88 & 80.79 & 82.13 & 83.08 & 83.74 & 84.36 & 84.70 & 85.02 & 85.39 & \textit{83.01} & $^{+0.07}$\\

        \hline

        CL$_\text{score}$ \cite{durand_learning_2019} & 77.59 & 81.68 & 83.07 & 84.05 & 84.41 & 84.76 & 84.93 & 85.18 & 85.39 & \textit{83.45} \\
        \quad+$\text{CFT}_\text{BP-ASL}$ & \textbf{78.41} & \textbf{81.91} & \textbf{83.23} & \textbf{84.17} & \textbf{84.55} & \textbf{84.91} & \textbf{85.10} & \textbf{85.34} & \textbf{85.55} & \textit{\textbf{83.69}} & $^{+0.24}$\\
        \quad+$\text{CFT}_\text{GA-AP}$ & 78.22 & 81.81 & 83.16 & 84.09 & 84.45 & 84.79 & 84.95 & 85.21 & 85.41 & \textit{83.57} & $^{+0.12}$\\
        \hline

        CL$_\text{score+}$ \cite{durand_learning_2019} & 77.77 & 81.63 & 83.16 & 83.98 & 84.39 & 84.70 & 84.95 & 85.16 & 85.36 & \textit{83.46} & \\
        \quad+$\text{CFT}_\text{BP-ASL}$ & \textbf{78.61} & \textbf{81.86} & \textbf{83.30} & \textbf{84.10} & \textbf{84.52} & \textbf{84.85} & \textbf{85.12} & \textbf{85.33} & \textbf{85.52} & \textit{\textbf{83.69}} & $^{+0.23}$\\
        \quad+$\text{CFT}_\text{GA-AP}$ & 78.35 & 81.79 & 83.23 & 84.03 & 84.42 & 84.72 & 84.97 & 85.18 & 85.38 & \textit{83.56} & $^{+0.10}$\\

        \hline
         
    \end{tabular}
    
\end{table*}

The 5 methods use the following training configurations unless stated otherwise.
ResNet-101 \cite{he_deep_2016} pre-trained on ImageNet-1K \cite{deng_imagenet_2009} is used as the model backbone.
Images are resized to $448^2$ and rescaled to $[0, 1]$.
Training images are augmented by RandAugment \cite{cubuk_randaugment_2020}.
Adam optimizer\cite{kingma_adam_2014} with initial learning rate $2\times10^{-4}$ scheduled by 1cycle policy \cite{smith_super-convergence_2019}, ASL ($\gamma^-=4, \gamma^+=0, m=0.05$) \cite{ridnik_asymmetric_2021}, true weight decay $1\times 10^{-4}$, and batch size 128, are used to update parameters for 60 epochs.
The exponential moving averages of the parameters are maintained.
For LL-R, we set $\Delta_{rel}=1$, batch size 16, and Initial learning rate $1\times 10^{-5}$.
For Selective, we set $\gamma^+=0, \gamma^-=1, \gamma^u=4, K=5,$ and $\eta=0.05$.
For CL$_\text{score}$ and CL$_\text{score+}$, we set $\theta=2$.

\subsubsection{Ablation Studies} \label{sec:mscoco_ablation}
We respectively apply $\text{CFT}_\text{BP-ASL}$ and $\text{CFT}_\text{GA-AP}$ to the 5 trained models.

For $\text{CFT}_\text{BP-ASL}$, we set $\gamma^+=0, \gamma^-=4, m=0.05$. Full-batch Adam optimizer with learning rate $1\times10^{-4}$ is used for updating parameters for 5000 epochs. For $\text{CFT}_\text{GA-AP}$, 50 individuals are evolved for 5000 generations. All individuals are initialized by encoding the original parameters. At each generation, the individual with the best fitness is transferred to the next generation. The parents are selected using roulette wheel selection.
The crossover operation randomly switches 20\% of the positions of two parents to produce offspring.
Each offspring has a 50\% chance of mutation by adding a random scalar between $[-0.001, 0.001]$ to each position.

The results are shown in \autoref{tab:mscoco_ablation}.
CFT effectively improves the mAP of the 5 trained models under all known label proportions.
In particular, CFT achieves more AP improvements under lower known label proportions, because more unknown labels generally imply more incorrect pseudo-labels harming the model performance.
CFT achieves higher mAP improvements to AN than other models, probably because AN simply assigns negative pseudo-labels to all unknown labels, which may produce more incorrect pseudo-labels than other methods.
$\text{CFT}_\text{BP-ASL}$ generally achieves more significant improvements than $\text{CFT}_\text{GA-AP}$, which is opposite to the results on the CheXpert dataset. 
We have discussed the results in Sect. \ref{sec:discuss_bp_ga}.

\subsubsection{Comparison to State-of-the-art} \label{sec:coco_sota}

We compare our best three models Selective+$\text{CFT}_\text{BP-ASL}$, CL$_\text{score}$+$\text{CFT}_\text{BP-ASL}$, and CL$_\text{score+}$+$\text{CFT}_\text{BP-ASL}$ to the state-of-the-art methods:
\underline{SSGRL} \cite{chen_learning_2019}, \underline{GCN-ML} \cite{chen_multi-label_2019}, \underline{KGGR} \cite{chen_knowledge-guided_2020}, and \underline{P-GCN} \cite{chen_learning_2021} are state-of-the-art methods in fully labeled MLC that use graph neural networks to model label dependencies.
Chen \etal \cite{chen_heterogeneous_2022} adapted these methods for partially labeled MLC by substituting the loss with partial-BCE loss \cite{durand_learning_2019} and reported the results.
\underline{Curriculum Labelling} \cite{durand_learning_2019} self-trains the model by gradually generating pseudo-labels of unknown labels.
\underline{Partial-BCE} \cite{durand_learning_2019} is a new BCE loss that ignores the contributions of unknown labels, with gradient magnitude adjustment based on label proportion.
\underline{Asymmetric loss} \cite{ridnik_asymmetric_2021} decouples the focusing parameter of the focal loss \cite{lin_focal_2017} to handle the positive-negative sample imbalance better.
\underline{SST} \cite{chen_structured_2022} and \underline{HST} \cite{chen_heterogeneous_2022} explore within-image and cross-image semantic correlations to help generate pseudo-labels of unknown labels.
\underline{SARB} \cite{pu_semantic-aware_2022} and \underline{DSRB} \cite{pu_dual-perspective_2023} exploit instance-level and prototype-level semantic representations to complement unknown labels.
\underline{SR} is a self-trained framework based on saliency regularization, plus consistency regularization to complement unknown labels, and train a model in supervised and unsupervised manners.
\underline{PU-MLC} \cite{yuan_positive_2023} only uses positive labels for model training based on Positive-Unlabeled learning.

The comparison is shown in \autoref{tab:mscoco_state_of_the_art}.
The comparison is based on ResNet-101 pre-trained on ImageNet-1K and image size $448^2$.
Our Selective+$\text{CFT}_\text{BP-ASL}$, CL$_\text{score}$+$\text{CFT}_\text{BP-ASL}$, and CL$_\text{score+}$+$\text{CFT}_\text{BP-ASL}$ respectively achieves average mAP 83.69\%, 83.69\% and 83.16\%, outperforming the previous best PU-MLC (81.19\%) and other state-of-the-art methods.

\begin{table*}[]
    \caption{Comparison to State-of-the-art on MS-COCO. In mAP\%($\uparrow$). (\textbf{Best} is in bold, \underline{second-best} is underlined.)}
    \label{tab:mscoco_state_of_the_art}
    \centering
    \begin{tabular}{llcccccccccc}
        \hline
        \multirow{2}{*}{Method} & \multirow{2}{*}{Published by} & \multicolumn{9}{c}{Known Label Proportion} & \multirow{2}{*}{\textit{Average}}\\
        \cline{3-11}
        & & 10\% & 20\% & 30\% & 40\% & 50\% & 60\% & 70\% & 80\% & 90\% & \\
        \hline

        SSGRL \cite{chen_learning_2019} & ICCV'19 & 62.50 & 70.50 & 73.20 & 74.50 & 76.30 & 76.50 & 77.10 & 77.90 & 78.40 & \textit{74.10} \\
        GCN-ML \cite{chen_multi-label_2019} & CVPR'19 & 63.80 & 70.90 & 72.80 & 74.00 & 76.70 & 77.10 & 77.30 & 78.30 & 78.60 & \textit{74.39} \\
        KGGR \cite{chen_knowledge-guided_2020} & TPAMI'22 & 66.60 & 71.40 & 73.80 & 76.70 & 77.50 & 77.90 & 78.40 & 78.70 & 79.10 & \textit{75.57} \\
        P-GCN \cite{chen_learning_2021} & TPAMI'21 & 67.50 & 71.60 & 73.80 & 75.50 & 77.40 & 78.40 & 79.50 & 80.70 & 81.50 & \textit{76.21} \\
        Curriculum Labelling \cite{durand_learning_2019} & CVPR'19 & 26.70 & 31.80 & 51.50 & 65.40 & 70.00 & 71.90 & 74.00 & 77.40 & 78.00 & \textit{60.74} \\
        Partial-BCE \cite{durand_learning_2019} & CVPR'19 & 61.60 & 70.50 & 74.10 & 76.30 & 77.20 & 77.70 & 78.20 & 78.40 & 78.50 & \textit{74.72} \\
        Asymmetric Loss \cite{ridnik_asymmetric_2021} & CVPR'21 & 69.70 & 74.00 & 75.10 & 76.80 & 77.50 & 78.10 & 78.70 & 79.10 & 79.70 & \textit{76.52} \\
        SST \cite{chen_structured_2022} & AAAI'22& 68.10 & 73.50 & 75.90 & 77.30 & 78.10 & 78.90 & 79.20 & 79.60 & 79.90 & \textit{76.72} \\
        HST \cite{chen_heterogeneous_2022} & arxiv'23 & 70.60 & 75.80 & 77.30 & 78.30 & 79.00 & 79.40 & 79.90 & 80.20 & 80.40 & \textit{77.88} \\
        SARB \cite{pu_semantic-aware_2022} & AAAI'22 & 71.20 & 75.00 & 77.10 & 78.30 & 78.90 & 79.60 & 79.80 & 80.50 & 80.50 & \textit{77.90} \\
        DSRB \cite{pu_dual-perspective_2023} & arxiv'23 & 72.50 & 76.00 & 77.60 & 78.70 & 79.60 & 79.80 & 80.00 & 80.50 & 80.80 & \textit{78.39} \\
        SR \cite{wang_saliency_2023} & CVPR'23 & 77.20 & 79.20 & 80.30 & 80.90 & 81.80 & 82.10 & 82.10 & 82.60 & 82.70 & \textit{81.00}\\
        PU-MLC \cite{yuan_positive_2023} & arxiv'23 & 75.70 & 78.60 & 80.20 & 81.30 & 82.00 & 82.60 & 83.00 & 83.50 & 83.80 & \textit{81.19} \\
        \textbf{(Ours) }Selective+$\text{CFT}_\text{BP-ASL}$ &  & 78.02 & 80.91 & 82.35 & 83.30 & 83.90 & 84.50 & 84.84 & 85.12 & 85.49 & \underline{\textit{83.16}} \\
        \textbf{(Ours) }CL$_\text{score}$+$\text{CFT}_\text{BP-ASL}$ & & \underline{78.41} & \textbf{81.91} & \underline{83.23} & \textbf{84.17} & \textbf{84.55} & \textbf{84.91} & \underline{85.10} & \textbf{85.34} & \textbf{85.55} & \textit{\textbf{83.69}} \\
        \textbf{(Ours) }CL$_\text{score+}$+$\text{CFT}_\text{BP-ASL}$ & & \textbf{78.61} & \underline{81.86} & \textbf{83.30} & \underline{84.10} & \underline{84.52} & \underline{84.85} & \textbf{85.12} & \underline{85.33} & \underline{85.52} & \textit{\textbf{83.69}} \\
        \hline
        
    \end{tabular}
\end{table*}

\subsubsection{Impact of the Feature Vector's Dimension} \label{sec:mscoco_z}
The feature vector's dimension $Z$ determines the number of parameters of each LR ($\text{\#params}=Z+1$). In this part, we study the effectiveness of CFT under different $Z$.
To modify $Z$, we add an fully-connected layer on the top of the backbone before model training. For example, if we want to change $Z$ from 2048 (the original $Z$ for ResNet-101) to 256, we add a 256-unit fully-connected layer.

\autoref{tab:dimension} reports the results of the model trained with AN. Both $\text{CFT}_\text{BP-ASL}$ and $\text{CFT}_\text{GA-AP}$ effectively improve the mAP under all tested $Z$. We also find that adding the layer can improve the average mAP of the trained model (before applying any CFT) to over 78\%, which is higher than not adding the layer (77.44\%).
Besides, as $Z$ becomes larger, the mAP improvements made by $\text{CFT}_\text{BP-ASL}$ and $\text{CFT}_\text{GA-AP}$ are also larger.
At $Z=16384$, AN+$\text{CFT}_\text{BP-ASL}$ achieves the best average mAP of 80.80\%.

\begin{table}[]
\caption{Results of different feature vector's dimension $Z$ on MS-COCO. In mAP\%($\uparrow$).
(\textcolor{red}{\textbf{Red Bold}} is the best across all $Z$)}
    \label{tab:dimension}
    \begin{tabular}{clcccc@{\hspace{-0.6\tabcolsep}}c}
        \hline
        \multirow{2}{*}{Dimension} & \multirow{2}{*}{Method} & \multicolumn{3}{c}{Known Label Prop.*} & \multirow{2}{*}{\textit{Average}} &\\
        \cline{3-5}
        & & 10\% & 50\% & 90\% & & \\
         \hline
        \multirow{3}{*}{\makecell{Original\\$Z=2048$}} & AN& 66.77 & 78.45 & 83.77 & \textit{77.44} \\
        & \quad+$\text{CFT}_\text{BP-ASL}$ & \textbf{70.83} & \textbf{79.82} & 83.83 & \textit{\textbf{78.88}} & $^{+1.44}$ \\
        & \quad+$\text{CFT}_\text{GA-AP}$ & 67.26 & 78.86 & \textbf{83.85} & \textit{77.77} & $^{+0.33}$\\
        \hline
        \multirow{3}{*}{\makecell{Added FC\\$Z=512$}} & AN & 67.98 & 79.07 & 83.99 & \textit{78.09}  \\
        & \quad+$\text{CFT}_\text{BP-ASL}$ & \textbf{70.57} & \textbf{80.19} & \textbf{84.17} & \textit{\textbf{79.15}} & $^{+1.06}$ \\
        & \quad+$\text{CFT}_\text{GA-AP}$ & 68.37 & 79.50 & 84.10 & \textit{78.44} & $^{+0.35}$ \\
        \hline
        \multirow{3}{*}{\makecell{Added FC\\$Z=1024$}} & AN & 68.30 & 79.25 & 84.19 & \textit{78.28} \\
        & \quad+$\text{CFT}_\text{BP-ASL}$ & \textbf{72.61} & \textbf{80.61} & \textbf{84.29} & \textbf{\textit{79.84}} & $^{+1.56}$ \\
        & \quad+$\text{CFT}_\text{GA-AP}$ & 69.09 & 79.74 & 84.26 & \textit{78.70} & $^{+0.42}$ \\
        \hline
        \multirow{3}{*}{\makecell{Added FC\\$Z=2048$}} & AN & 68.31 & 79.50 & 84.32 & \textit{78.38} \\
        & \quad+$\text{CFT}_\text{BP-ASL}$ & \textbf{73.42} & \textbf{80.79} & \textbf{84.43} & \textbf{\textit{80.21}} & $^{+1.84}$ \\
        & \quad+$\text{CFT}_\text{GA-AP}$ & 69.05 & 79.87 & 84.40 & \textit{78.82} & $^{+0.45}$ \\
        \hline
        \multirow{3}{*}{\makecell{Added FC\\$Z=4096$}} & AN & 68.82 & 79.64 & 84.38 & \textit{78.56} \\
        & \quad+$\text{CFT}_\text{BP-ASL}$ & \textbf{73.85} & \textbf{81.10} & \textbf{84.50} & \textbf{\textit{80.48}} & $^{+1.91}$ \\
        & \quad+$\text{CFT}_\text{GA-AP}$ & 69.64 & 80.05 & 84.45 & \textit{79.01} & $^{+0.45}$ \\
        \hline
        \multirow{3}{*}{\makecell{Added FC\\$Z=8192$}} & AN & 68.95 & 79.68 & 84.46 & \textit{78.65}\\
        & \quad+$\text{CFT}_\text{BP-ASL}$ & \textbf{74.30} & \textbf{81.42} & \textcolor{red}{\textbf{84.58}} & \textbf{\textit{80.71}} & $^{+2.06}$ \\
        & \quad+$\text{CFT}_\text{GA-AP}$ & 69.83 & 80.21 & 84.54 & \textit{79.15} & $^{+0.50}$ \\
        \hline
        \multirow{3}{*}{\makecell{Added FC\\$Z=16384$}} & AN & 68.95 & 79.55 & 84.35 & \textit{78.54} \\
        & \quad+$\text{CFT}_\text{BP-ASL}$ & \textbf{74.70} & \textcolor{red}{\textbf{81.49}} & \textbf{84.50} & \textcolor{red}{\textbf{\textit{80.80}}} & $^{+2.26}$ \\
        & \quad+$\text{CFT}_\text{GA-AP}$ & 70.02 & 80.06 & 84.40 & \textit{79.06} & $^{+0.52}$ \\
        \hline
        \multirow{3}{*}{\makecell{Added FC\\$Z=32768$}} & AN & 68.75 & 79.24 & 84.27 & \textit{78.32} \\
        & \quad+$\text{CFT}_\text{BP-ASL}$ & \textcolor{red}{\textbf{74.89}} & \textbf{81.38} & \textbf{84.48} & \textbf{\textit{80.78}} & $^{+2.47}$ \\
        & \quad+$\text{CFT}_\text{GA-AP}$ & 69.90 & 79.64 & 84.33 & \textit{78.81} & $^{+0.50}$ \\
        \hline
    \end{tabular}
    \vspace{-2pt}
    
    \footnotesize{~* Results under 20\%, 30\%, 40\%, 60\%, 70\%, and 80\% are omitted.}
\end{table}

\subsubsection{Effectiveness on Different Model Architectures} \label{sec:mscoco_backbone}
We also evaluate the effectiveness of CFT on different model architectures, including Vision Transformer (ViT) \cite{dosovitskiy_image_2020}, SwinTransformer-v2 \cite{liu_swin_2022}, EfficientNet-v2 \cite{tan_efficientnetv2_2021}, and ConvNext-v2 \cite{woo_convnext_2023}. The results are shown in \autoref{tab:model}. Our methods effectively improve the mAP on all tested architectures.

\begin{table}[]
    \caption{Results on various architectures on MS-COCO. In mAP\%($\uparrow$).}
    \label{tab:model}
    \centering
    \begin{tabular}{llccc}
        \hline
        \multirow{2}{*}{Model Architecture} & \multirow{2}{*}{Method} & \multicolumn{3}{c}{Known Label Prop.}\\
        \cline{3-5}
        & & 10\% & 50\% & 90\% \\
        
        \hline

        \multirow{2}{*}{ResNet-101 \cite{he_deep_2016}} & AN & 68.95 & 79.55 & 84.35 \\
        & \quad+$\text{CFT}_\text{BP-ASL}$ & \textbf{74.70} & \textbf{81.49} & \textbf{84.50} \\
        & \quad+$\text{CFT}_\text{GA-AP}$ & 70.02 & 80.06 & 84.40 \\
        \hline

        \multirow{2}{*}{ViT-L/16 \cite{dosovitskiy_image_2020}} & AN & 56.11 & 70.32 & 75.24 \\
        & \quad+$\text{CFT}_\text{BP-ASL}$ & \textbf{63.56} & \textbf{72.20} & \textbf{75.47} \\
        & \quad+$\text{CFT}_\text{GA-AP}$ & 56.62 & 70.45 & 75.29 \\
        \hline

        \multirow{2}{*}{SwinTrans-v2-B \cite{liu_swin_2022}} & AN & 69.18 & 78.21 & 82.07 \\
        & \quad+$\text{CFT}_\text{BP-ASL}$ & \textbf{72.73} & \textbf{79.24} & \textbf{82.13} \\
        & \quad+$\text{CFT}_\text{GA-AP}$ &69.90 & 78.47 & 82.12\\
        \hline

        \multirow{2}{*}{EfficientNetV2-L \cite{tan_efficientnetv2_2021}} & AN & 73.85 & 83.18 & 86.79 \\
        & \quad+$\text{CFT}_\text{BP-ASL}$ & \textbf{78.94} & \textbf{84.82} & \textbf{87.09} \\
        & \quad+$\text{CFT}_\text{GA-AP}$ & 74.48 & 83.57 & 86.93 \\
        \hline

        \multirow{2}{*}{ConvNextV2-L \cite{woo_convnext_2023}} & AN & 78.37 & 87.57 & 90.31 \\
        & \quad+$\text{CFT}_\text{BP-ASL}$ & \textbf{84.32} & \textbf{89.16} & \textbf{90.53} \\
        & \quad+$\text{CFT}_\text{GA-AP}$ & 78.79 & 87.71 & 90.34\\
        \hline
        
    \end{tabular}
\end{table}

\subsection{Open Images V3}\label{sec:oiv3}
\subsubsection{Dataset}
Open Images V3 \cite{kuznetsova_open_2020} is a real partially-labeled dataset with 5000 trainable categories. The training, validation, and test sets are all partially labeled, consisting of 9 million, 41,620, and 125,436 images, respectively. In the training set, only 0.086\% of the labels are known. This sparsity level of known labels makes training high-performance models more challenging. Following previous work on this dataset, the training and validation sets are used for model training and tuning hyperparameters, respectively. We report the results on the test set.

\subsubsection{Performance Metric}

The model performance is evaluated by the mean of per-category AP (mAP) of all 5000 categories.
Additionally, to better compare to other methods, we follow previous work to sort the 5000 categories ascendingly by their number of known labels, and evenly divide the sorted categories into five groups. Group 1 (G1) consists of the 1000 categories with the fewest known labels, and G5 consists of the 1000 categories with the most known labels. The per-group mAP is also reported.

\subsubsection{Model Training}

Same as in the experiments on the MS-COCO dataset in Sect. \ref{sec:mscoco_training}, we train 5 models with AN, LL-R, Selective, CL$_\text{score}$, and CL$_\text{score+}$, respectively.

The training configurations are the same as in the experiments on the MS-COCO dataset (Sect. \ref{sec:mscoco_training}) except the followings.
Images are resized to $224^2$. Batch size is set to 256. For LL-R, we set $\Delta_{rel}=0.01$. For Selective, we set $\gamma^+=1, \gamma^-=2, \gamma^u=7, K=50,$ and $\eta=0.05$.

\subsubsection{Ablation Studies}

\begin{table}
\caption{Ablation studies on Open Images V3 test set. In mAP\%($\uparrow$).}
    \label{tab:oiv3_result}
    \centering
    \begin{tabular}{lcccccc}
         \hline
         Method & G1 & G2 & G3 & G4 & G5 & \textit{All}\\
         \hline
         AN & 78.71 & 81.59 & 84.02 & 86.18 & 90.60 & \textit{84.28}\\
         \quad+$\text{CFT}_\text{BP-ASL}$ & \textbf{79.39} & \textbf{82.20} & \textbf{84.85} & \textbf{87.32} & \textbf{91.89} & \textbf{\textit{85.19}}\\
         \quad+$\text{CFT}_\text{GA-AP}$ & 79.09 & 82.02 & 84.69 & 87.07 & 91.70 & \textit{84.97} \\
         \hline
         LL-R \cite{kim_large_2022} & 78.89 & 81.68 & 84.17 & 86.38 & 90.95 & \textit{84.47} \\
         \quad+$\text{CFT}_\text{BP-ASL}$ & \textbf{79.30} & \textbf{82.32} & \textbf{85.01} & \textbf{87.46} & \textbf{92.12} & \textbf{\textit{85.31}}\\
         \quad+$\text{CFT}_\text{GA-AP}$ & 79.16 & 82.13 & 84.73 & 87.22 & 91.80 & \textit{85.07} \\
         \hline
         Selective \cite{ben-baruch_multi-label_2022} & 78.12 & 81.17 & 84.49 & 87.21 & 92.13 & \textit{84.69}\\
         \quad+$\text{CFT}_\text{BP-ASL}$ & \textbf{78.39} & \textbf{81.40} & \textbf{84.78} & \textbf{87.50} & \textbf{92.40} & \textbf{\textit{84.96}} \\
         \quad+$\text{CFT}_\text{GA-AP}$ & 78.20 & 81.18 & 84.62 & 87.37 & 92.24 & \textit{84.79} \\
         \hline
         CL$_\text{score}$ \cite{durand_learning_2019} & 77.45 & 80.59 & 84.18 & 86.76 & 91.90 & \textit{84.25} \\
         \quad+$\text{CFT}_\text{BP-ASL}$ & \textbf{77.82} & \textbf{80.85} & \textbf{84.43} & \textbf{87.05} & \textbf{92.12} & \textbf{\textit{84.52}} \\
         \quad+$\text{CFT}_\text{GA-AP}$ & 77.52 & 80.50 & 84.15 & 86.89 & 92.00 & \textit{84.28} \\
         \hline
         CL$_\text{score+}$ \cite{durand_learning_2019} & 77.61 & 80.47 & 84.10 & 86.80 & 91.98 & \textit{84.26} \\
         \quad+$\text{CFT}_\text{BP-ASL}$ & \textbf{77.86} & \textbf{80.81} & \textbf{84.34} & \textbf{87.12} & \textbf{92.19} & \textbf{\textit{84.54}} \\
         \quad+$\text{CFT}_\text{GA-AP}$ & 77.63 & 80.48 & 84.13 & 86.96 & 92.10 & \textit{84.33} \\
         \hline
    \end{tabular}
\end{table}

\begin{figure}
    \centering
    \includegraphics[width=0.49\linewidth]{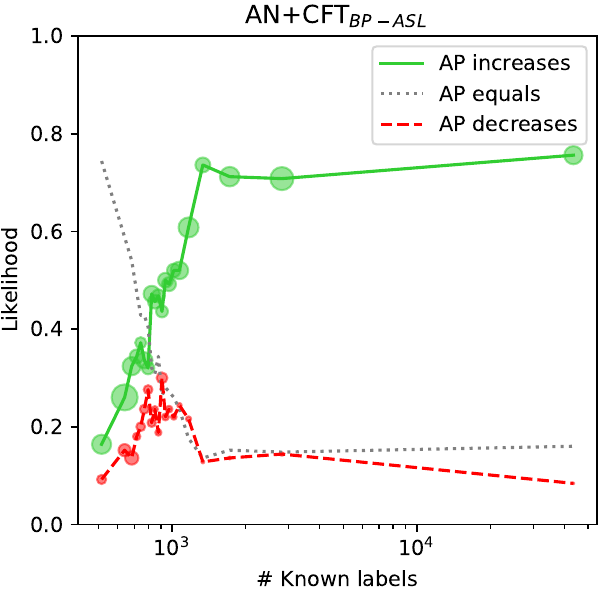}
    \hfill
    \includegraphics[width=0.49\linewidth]{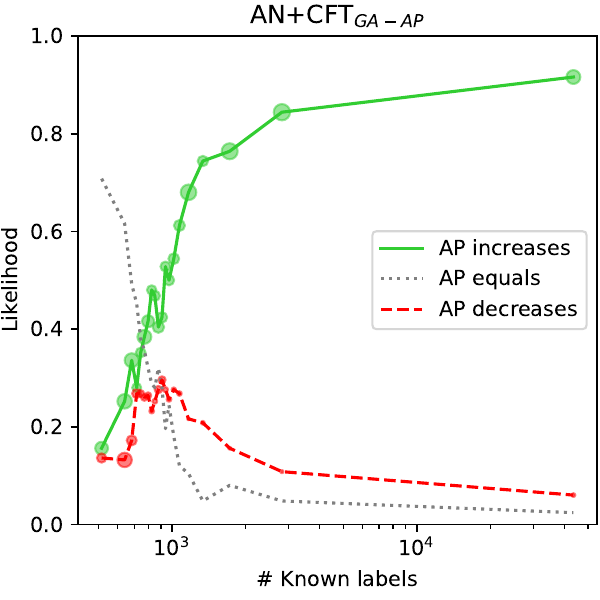}

    \caption{The likelihood (y-axis) of categories with different numbers of known labels (x-axis) getting AP increased/decreased/unchanged from (\textbf{left}) CFT$_\text{BP-ASL}$ and (\textbf{right}) CFT$_\text{GA-AP}$. The scatters' radius represent the mean AP changes.}
    \label{fig:oiv3_likelihood}
\end{figure}

We respectively apply $\text{CFT}_\text{BP-ASL}$ and $\text{CFT}_\text{GA-AP}$ to the three trained models. Their configurations are the same as on the MS-COCO dataset (Sect. \ref{sec:mscoco_ablation}).

The results in \autoref{tab:oiv3_result} show that both $\text{CFT}_\text{BP-ASL}$ and $\text{CFT}_\text{GA-AP}$ can effectively improve the mAP of the 5 trained models. $\text{CFT}_\text{BP-ASL}$ achieves larger improvements than $\text{CFT}_\text{GA-AP}$. We have discussed the results in Sect. \ref{sec:discuss_bp_ga}.

To further analyze the results, we focus on the result of AN+CFT$_\text{BP-ASL}$ and assess the per-category AP improvements.
After applying CFT$_\text{BP-ASL}$ to AN, 2,376 categories get AP increased by mean +2.29\%, 1,666 categories get AP unchanged, and 958 categories get AP decreased by mean -1.00\%.
In particular, we found that these per-category AP changes are related to their number of known labels.
Specifically, we sort all 5000 categories ascendingly by their number of known labels and equally divide them into 20 splits.
For each split, we calculate the average number of known labels, the likelihood of a category's AP being increased/decreased/unchanged, as well as the mean AP increases/decreases.
For example, in the $1^\text{st}$ split that consists of the 250 categories with the fewest known labels, the average number of known labels is 515. 41/23/186 categories get their AP increased/decreased/unchanged. Among them, the 41/23 categories get AP increased/decreased by mean +1.78\%/-1.29\%, respectively.
With this information, we can estimate the likelihood of a category with 515 known labels getting AP increased/decreased/unchanged is 16.4\%, 9.2\%, and 74.4\%, respectively, and the expected AP increases/decreases is +1.78\%/-1.29\%.
By counting this information of all 20 splits, we obtain a figure (\autoref{fig:oiv3_likelihood} left) showing, as the number of known labels changes, how the likelihoods of AP increased/decreased/unchanged and the magnitudes change.

As can be seen, if a category has very few known labels (\eg, $\le$1000), it is difficult to predict if its AP gets increased/decreased/unchanged after applying $\text{CFT}_\text{BP-ASL}$.
We surmise that the lack of known labels leads to the lack of training and validation samples for CFT$_\text{BP-ASL}$ to fine-tune the corresponding LR, which may respectively lead to poor decision boundary \cite{ben-baruch_multi-label_2022} and difficulty in precise evaluation of the performance. Hence, the LR may not be fine-tuned well.
On the other hand, if a category has sufficient known labels (\eg, $\ge$2000), it has a high likelihood of increasing AP and a low likelihood of decreasing/remaining unchanged. Even if it gets AP decreased, the magnitude of AP decrease is very likely to be negligible.

Overall, the total AP increases (5438.68\%) made by CFT$_\text{BP-ASL}$ greatly outweigh the total AP decreases (-960.18\%).
Hence, CFT is effective in improving the performance of the trained model.

\subsubsection{Comparison to State-of-the-art}
We compare our best three models Selective+$\text{CFT}_\text{BP-ASL}$, AN+$\text{CFT}_\text{BP-ASL}$, and LL-R+$\text{CFT}_\text{BP-ASL}$ to other state-of-the-art methods, including:
\underline{Curriculum Labelling} \cite{durand_learning_2019} and \underline{SR} have been introduced in Sect. \ref{sec:coco_sota}.
\underline{IMCL} \cite{huynh_interactive_2020} trains the model interactively with a similarity learner to leverage label and image dependencies.
\underline{LL-R}, \underline{LL-Ct}, and \underline{LL-Cp} \cite{kim_large_2022} identify false negative labels of AN based on loss magnitude and reject/temporarily correct/permanently correct them.
\underline{BoostLU} \cite{kim_bridging_2023} boosts the attribution scores of the highlighted regions in the class activation maps to compensate for the score decrease caused by false negative labels.

The comparison is shown in \autoref{tab:oiv3_sota}. The comparison is based on ResNet-101 pre-trained on ImageNet-1K and image size $224^2$. Our models achieve mAP 85.31\%, 85.19\%, and 84.96\%, outperforming the previous best SR (84.40\%) and other known state-of-the-art methods.

\begin{table*}[]
    
    \parbox{0.67\textwidth}{
        \caption{Comparison to State-of-the-art in Open Images V3 test set. In mAP\%($\uparrow$).}
        \label{tab:oiv3_sota}
        \centering
        \begin{tabular}{llcccccc}
             \hline
             Method & Published by & G1 & G2 & G3 & G4 & G5 & \textit{All}\\
             \hline
             Curriculum Labeling \cite{durand_learning_2019} & CVPR'19 & 70.37 & 71.32 & 76.23 & 80.54 & 86.81 & \textit{77.05}\\
             IMCL \cite{huynh_interactive_2020} & CVPR'20 & 70.95 & 72.59 & 77.64 & 81.83 & 87.34 & \textit{78.07}\\
             Selective \cite{ben-baruch_multi-label_2022} & CVPR'22 & 73.19 & 78.61 & \textbf{85.11} & \textbf{87.70} & 90.61 & \textit{83.03}\\
             LL-R \cite{kim_large_2022} & CVPR'22 & 77.76 & 79.07 & 81.94 & 84.51 & 89.36 & \textit{82.53}\\
             LL-Ct \cite{kim_large_2022} & CVPR'22 & 77.76 & 79.18 & 81.97 & 84.46 & 89.51 & \textit{82.58}\\
    
             LL-Cp \cite{kim_large_2022} & CVPR'22 & 77.49 & 79.22 & 81.89 & 84.51 & 89.18 & \textit{82.46}\\
             LL-R + BoostLU \cite{kim_bridging_2023} & CVPR'23 & 79.28 & 80.81 & 83.32 & 85.63 & 90.27 & \textit{83.86}\\
             LL-Ct + BoostLU \cite{kim_bridging_2023} & CVPR'23 & \underline{79.43} & 80.75 & 83.41 & 85.70 & 90.41 & \textit{83.94}\\
             LL-Cp + BoostLU \cite{kim_bridging_2023} & CVPR'23 & \textbf{79.53} & 81.04 & 83.40 & 85.85 & 90.39 & \textit{84.04}\\
             SR \cite{wang_saliency_2023} & ICCV'23 & 78.90 & 80.90 & 83.70 & 86.80 & 91.40 & \textit{84.40}\\
             \textbf{(Ours)} Selective+$\text{CFT}_\text{BP-ASL}$ & & 78.39 & 81.40 & 84.78 & \underline{87.50} & \textbf{92.40} & \textit{84.96} \\
             \textbf{(Ours)} AN+$\text{CFT}_\text{BP-ASL}$ & & 79.39 & \underline{82.20} & 84.85 & 87.32 & 91.89 & \textit{\underline{85.19}} \\
             \textbf{(Ours)} LL-R+$\text{CFT}_\text{BP-ASL}$ & & 79.30 & \textbf{82.32} & \underline{85.01} & 87.46 & \underline{92.12} & \textbf{\textit{85.31}}\\

             \hline
        \end{tabular}
    }
    ~
    \parbox{0.3\textwidth}{
            \caption{Computation time (in min.) in experiments. (O.I.V3 = Open Images V3)}
            \label{tab:dataset_computation_time}
            \centering
            \setlength{\tabcolsep}{5pt}
            \begin{tabular}{lrrr}
                \hline
                 Method & CheXpert & COCO & O.I.V3\\
                 \hline
                 $\text{CFT}_\text{BP-ASL}$ & 8.3 & 2.7 & 38.9\\
                 $\text{CFT}_\text{GA}$ & 73.5 & 15.1 & 1267.4\\
                 \hline
            \end{tabular}
            \vspace{20pt}
            \caption{Mean computation time with standard deviation.}
            \label{tab:computation_time}
            \centering
            \begin{tabular}{lr}
                \hline
                 Method & Per-LR Time (sec)\\
                 \hline
                 $\text{CFT}_\text{BP-ASL}$ & 1.07 $\pm$0.06\\
                 $\text{CFT}_\text{GA}$(\#p=10) & 4.17 $\pm$0.01\\
                 $\text{CFT}_\text{GA}$(\#p=50) & 17.65 $\pm$0.04\\
                 \hline
            \end{tabular}
        }
\end{table*}

\subsection{Discussion} \label{sec:discussion}

\subsubsection{Effectiveness and Generalizability of CFT}

By applying $\text{CFT}_\text{BP-ASL}$ or $\text{CFT}_\text{GA}$ to trained models with different methods generating pseudo-labels, we have achieved state-of-the-art results on the three datasets (\autoref{tab:dataset}) with different image types, scales of samples and categories, performance metrics, and known label proportions. CFT is also consistently effective in models with different backbones (Sect. \ref{sec:mscoco_backbone}) and feature vector's dimensions (Sect. \ref{sec:mscoco_z}). These outstanding results suggest that CFT is promising to be substantial and prevalent for developing deep classification models.

\subsubsection{$\text{CFT}_\text{BP-ASL}$ vs. $\text{CFT}_\text{GA}$} \label{sec:discuss_bp_ga}
$\text{CFT}_\text{GA}$ significantly outperforms $\text{CFT}_\text{BP-ASL}$ on the CheXpert dataset, while $\text{CFT}_\text{BP-ASL}$ is more effective than $\text{CFT}_\text{GA}$ on the MS-COCO and Open Images V3 datasets.
Due to the naturalness of the datasets and the lack of interpretability of deep learning, it is difficult to find the fundamental reasons. Hence, we suggest a few possible key factors of their effectiveness.\\
\textbf{Model's Performance Metric}. Observing the model performance metrics on the three datasets (\autoref{tab:dataset}). $\text{CFT}_\text{BP-ASL}$ is more effective to the datasets with mAP metric, whereas $\text{CFT}_\text{GA}$ is more effective to the dataset with mAUC metric.\\
\textbf{Image Type.} $\text{CFT}_\text{BP-ASL}$ is more effective to the natural image datasets (MS-COCO and Open Images V3), whereas $\text{CFT}_\text{GA}$ is more effective to the medical image dataset (CheXpert).
\textbf{Efficiency.} BP has been widely used to train neural networks in decade, which is undeniably more effective than GA.\\
\textbf{Direct Optimization}. GA can directly optimize non-differentiable metric such as AUC and AP. BP indirectly optimize them by optimizing surrogate losses (\eg, ASL) or their differentiable approximations.

\subsubsection{Limitations}
CFT does not consider the instance similarity and label correlation.
The model performance after applying CFT greatly depends on how well the model is trained (before applying CFT), as CFT only fine-tunes the model's classification layer, which is only a tiny part of the model architecture.
Addressing these limitations could serve as potential directions for enhancing the effectiveness of CFT in future research.

\subsection{Computation Time Evaluation} \label{sec:time_eval}

\autoref{tab:dataset_computation_time} reports the computation time of CFT$_\text{BP-ASL}$ and CFT$_\text{GA}$ in the experiments on the three datasets. However, these results are interfered with by various variables such as hardware, disk IO, and implementation details.

Hence, we systematically measure the computation time of $\text{CFT}_\text{BP-ASL}$ and $\text{CFT}_\text{GA}$ by running them on pseudo training samples under the same environments and variables.
Each pseudo training sample consists of a random float-32 feature vector $\mathbf{z}$ and random label vector $\mathbf{y}$. The experiments are conducted on a desktop computer with i9-13900kf CPU, RTX 4090 24G, and 128G RAM.

We first test the computation time under the following variables. The number of training samples $N=300000$, feature vector dimension $Z=512$, known label proportion 10\%, and 1000 epochs. For $\text{CFT}_\text{GA}$, we respectively test 50 and 10 individuals, denoted as \#p=50 and \#p=10. \#p=50 is used in the experiments in MS-COCO and Open Images V3.

We report per-LR computation time in \autoref{tab:computation_time}, $\text{CFT}_\text{GA}$ takes 3.9 to 16.5 times more than $\text{CFT}_\text{BP-ASL}$. More individuals requires more time, as GA needs to evaluate all individuals in each generation. Even though, the computation time of $\text{CFT}_\text{GA}$ is acceptable, as using $\text{CFT}_\text{GA}$ to fine-tune models trained on Open Images V3 (5,000 LRs) takes only less than 24 hours, as in \autoref{tab:dataset_computation_time}.

\begin{figure}
    \centering
    \includegraphics[width=\linewidth]{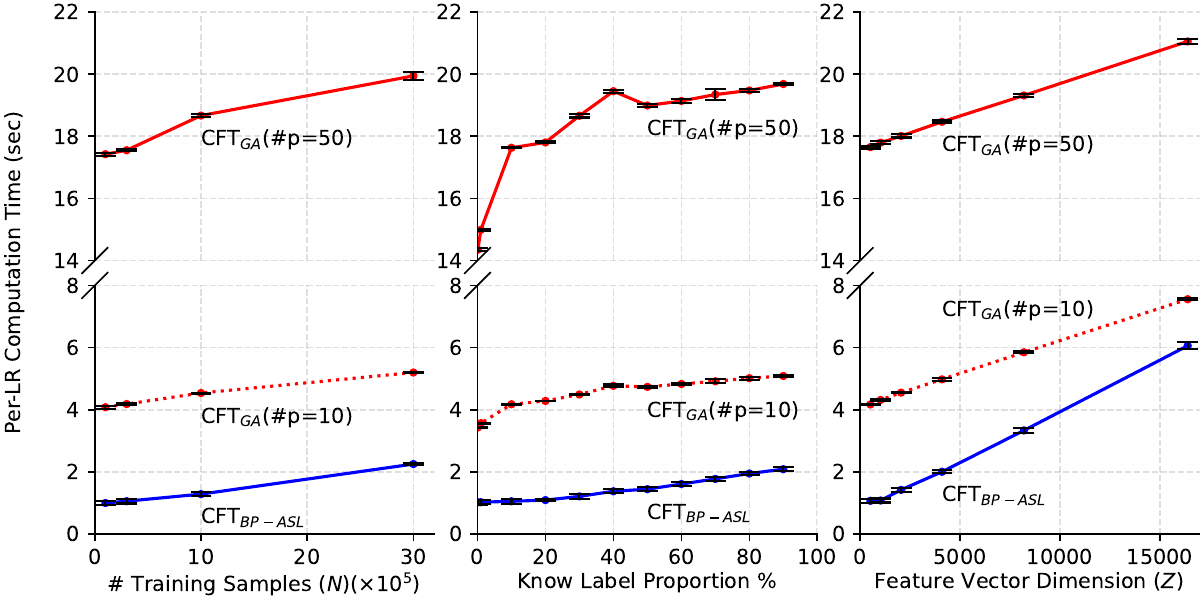}
    
    \caption{Per-LR computation time in seconds of different CFT varieties.}
    \label{fig:computation_cost}
\end{figure}

Subsequently, we vary (1) $N$, (2) known label proportions, and (3) $Z$ to evaluate their impacts on the computation time. \autoref{fig:computation_cost} reports the per-AP computation time. The number of training samples is not proportional to the computation time. For example, in $\text{CFT}_\text{BP-ASL}$, $N=100000$ takes 1.00 sec/LR while $N=3000000$ takes 2.25 sec/LR. 30x increase of $N$ only increases the computation time by 2.25x.
A similar phenomenon can also be observed in the changes in the known label proportion.
These results show that our methods are scalable to datasets with large numbers of training samples and known label proportions.

\section{Conclusion}\label{sec:conclusion}

This paper proposes a new method named Category-wise Fine-Tuning (CFT) that fine-tunes trained deep classification models to mitigate the incorrectness learned from incorrect pseudo-labels.
The two CFT varieties, CFT$_\text{BP-ASL}$ and CFT$_\text{GA}$, the former fine-tunes LRs by minimizing the asymmetric losses using backpropagation, and the latter fine-tunes LRs by directly maximizing classification performances using Genetic Algorithm, an evolutionary algorithm seldom used for training deep models.
Comprehensive experiments on CheXpert, MS-COCO, and Open Images V3 show the remarkable effectiveness and generalizability of CFT.
In particular, we achieve state-of-the-art results on all three datasets we used. CFT significantly improves the classification performances of models trained with various methods, including Assume Negative, LL-R, Selective, and Curriculum Labeling.
Moreover, CFT is also effective for trained models with different architectures (various CNNs and Transformers) and feature vector's dimensions. Last but not least, the computation time of CFT is systematically evaluated, revealing an acceptable short processing time. This characteristic renders CFT easily integrable as an add-on to the current training process. These outstanding experimental results suggest that the proposed CFT is promising to be a substantial and prevalent method for developing deep classification models on partially labeled datasets.

\section*{Acknowledgments}
This work is supported by Macao Polytechnic University under grant number RP/FCA-04/2022.

 
\bibliography{IEEEabrv, references}

\begin{thebibliography}{10}
\providecommand{\url}[1]{#1}
\csname url@samestyle\endcsname
\providecommand{\newblock}{\relax}
\providecommand{\bibinfo}[2]{#2}
\providecommand{\BIBentrySTDinterwordspacing}{\spaceskip=0pt\relax}
\providecommand{\BIBentryALTinterwordstretchfactor}{4}
\providecommand{\BIBentryALTinterwordspacing}{\spaceskip=\fontdimen2\font plus
\BIBentryALTinterwordstretchfactor\fontdimen3\font minus \fontdimen4\font\relax}
\providecommand{\BIBforeignlanguage}[2]{{%
\expandafter\ifx\csname l@#1\endcsname\relax
\typeout{** WARNING: IEEEtran.bst: No hyphenation pattern has been}%
\typeout{** loaded for the language `#1'. Using the pattern for}%
\typeout{** the default language instead.}%
\else
\language=\csname l@#1\endcsname
\fi
#2}}
\providecommand{\BIBdecl}{\relax}
\BIBdecl

\bibitem{irvin_chexpert_2019}
J.~Irvin, P.~Rajpurkar, M.~Ko, Y.~Yu, S.~Ciurea-Ilcus, C.~Chute, H.~Marklund, B.~Haghgoo, R.~Ball, K.~Shpanskaya, and {others}, ``Chexpert: {A} large chest radiograph dataset with uncertainty labels and expert comparison,'' in \emph{Proceedings of the {AAAI} conference on artificial intelligence}, vol.~33, 2019, pp. 590--597.

\bibitem{huynh_interactive_2020}
D.~Huynh and E.~Elhamifar, ``Interactive multi-label cnn learning with partial labels,'' in \emph{Proceedings of the {IEEE}/{CVF} {Conference} on {Computer} {Vision} and {Pattern} {Recognition}}, 2020, pp. 9423--9432.

\bibitem{lin_rethinking_2022}
J.~Lin, T.~Yu, and Z.~J. Wang, ``Rethinking {Crowdsourcing} {Annotation}: {Partial} {Annotation} {With} {Salient} {Labels} for {Multilabel} {Aerial} {Image} {Classification},'' \emph{IEEE transactions on geoscience and remote sensing}, vol.~60, pp. 1--12, 2022, publisher: IEEE.

\bibitem{ridnik_asymmetric_2021}
T.~Ridnik, E.~Ben-Baruch, N.~Zamir, A.~Noy, I.~Friedman, M.~Protter, and L.~Zelnik-Manor, ``Asymmetric loss for multi-label classification,'' in \emph{Proceedings of the {IEEE}/{CVF} {International} {Conference} on {Computer} {Vision}}, 2021, pp. 82--91.

\bibitem{kobayashi_two-way_2023}
T.~Kobayashi, ``Two-{Way} {Multi}-{Label} {Loss},'' in \emph{Proceedings of the {IEEE}/{CVF} {Conference} on {Computer} {Vision} and {Pattern} {Recognition}}, 2023, pp. 7476--7485.

\bibitem{chen_learning_2019}
T.~Chen, M.~Xu, X.~Hui, H.~Wu, and L.~Lin, ``Learning semantic-specific graph representation for multi-label image recognition,'' in \emph{Proceedings of the {IEEE}/{CVF} international conference on computer vision}, 2019, pp. 522--531.

\bibitem{durand_learning_2019}
T.~Durand, N.~Mehrasa, and G.~Mori, ``Learning a deep convnet for multi-label classification with partial labels,'' in \emph{Proceedings of the {IEEE}/{CVF} conference on computer vision and pattern recognition}, 2019, pp. 647--657.

\bibitem{sun_dualcoop_2022}
X.~Sun, P.~Hu, and K.~Saenko, ``Dualcoop: {Fast} adaptation to multi-label recognition with limited annotations,'' \emph{arXiv preprint arXiv:2206.09541}, 2022.

\bibitem{hu_dualcoop_2023}
P.~Hu, X.~Sun, S.~Sclaroff, and K.~Saenko, ``{DualCoOp}++: {Fast} and {Effective} {Adaptation} to {Multi}-{Label} {Recognition} with {Limited} {Annotations},'' \emph{arXiv preprint arXiv:2308.01890}, 2023.

\bibitem{ding_exploring_2023}
Z.~Ding, A.~Wang, H.~Chen, Q.~Zhang, P.~Liu, Y.~Bao, W.~Yan, and J.~Han, ``Exploring {Structured} {Semantic} {Prior} for {Multi} {Label} {Recognition} with {Incomplete} {Labels},'' in \emph{Proceedings of the {IEEE}/{CVF} {Conference} on {Computer} {Vision} and {Pattern} {Recognition}}, 2023, pp. 3398--3407.

\bibitem{kim_large_2022}
Y.~Kim, J.~M. Kim, Z.~Akata, and J.~Lee, ``Large loss matters in weakly supervised multi-label classification,'' in \emph{Proceedings of the {IEEE}/{CVF} {Conference} on {Computer} {Vision} and {Pattern} {Recognition}}, 2022, pp. 14\,156--14\,165.

\bibitem{kundu_exploiting_2020}
K.~Kundu and J.~Tighe, ``Exploiting weakly supervised visual patterns to learn from partial annotations,'' \emph{Advances in Neural Information Processing Systems}, vol.~33, pp. 561--572, 2020.

\bibitem{bucak_multi-label_2011}
S.~S. Bucak, R.~Jin, and A.~K. Jain, ``Multi-label learning with incomplete class assignments,'' in \emph{{CVPR} 2011}.\hskip 1em plus 0.5em minus 0.4em\relax IEEE, 2011, pp. 2801--2808.

\bibitem{chen_fast_2013}
M.~Chen, A.~Zheng, and K.~Weinberger, ``Fast image tagging,'' in \emph{International conference on machine learning}.\hskip 1em plus 0.5em minus 0.4em\relax PMLR, 2013, pp. 1274--1282.

\bibitem{lin_microsoft_2014}
T.-Y. Lin, M.~Maire, S.~Belongie, J.~Hays, P.~Perona, D.~Ramanan, P.~Dollár, and C.~L. Zitnick, ``Microsoft coco: {Common} objects in context,'' in \emph{European conference on computer vision}.\hskip 1em plus 0.5em minus 0.4em\relax Springer, 2014, pp. 740--755.

\bibitem{ben-baruch_multi-label_2022}
E.~Ben-Baruch, T.~Ridnik, I.~Friedman, A.~Ben-Cohen, N.~Zamir, A.~Noy, and L.~Zelnik-Manor, ``Multi-label classification with partial annotations using class-aware selective loss,'' in \emph{Proceedings of the {IEEE}/{CVF} {Conference} on {Computer} {Vision} and {Pattern} {Recognition}}, 2022, pp. 4764--4772.

\bibitem{chen_structured_2022}
T.~Chen, T.~Pu, H.~Wu, Y.~Xie, and L.~Lin, ``Structured semantic transfer for multi-label recognition with partial labels,'' in \emph{Proceedings of the {AAAI} conference on artificial intelligence}, vol.~36, 2022, pp. 339--346, issue: 1.

\bibitem{arazo_pseudo-labeling_2020}
E.~Arazo, D.~Ortego, P.~Albert, N.~E. O’Connor, and K.~McGuinness, ``Pseudo-labeling and confirmation bias in deep semi-supervised learning,'' in \emph{2020 {International} {Joint} {Conference} on {Neural} {Networks} ({IJCNN})}.\hskip 1em plus 0.5em minus 0.4em\relax IEEE, 2020, pp. 1--8.

\bibitem{zhang_enhanced_2016}
Z.~Zhang, F.~Ringeval, B.~Dong, E.~Coutinho, E.~Marchi, and B.~Schüller, ``Enhanced semi-supervised learning for multimodal emotion recognition,'' in \emph{2016 {IEEE} international conference on acoustics, speech and signal processing ({ICASSP})}.\hskip 1em plus 0.5em minus 0.4em\relax IEEE, 2016, pp. 5185--5189.

\bibitem{mitchell_introduction_1998}
M.~Mitchell, \emph{An introduction to genetic algorithms}.\hskip 1em plus 0.5em minus 0.4em\relax MIT press, 1998.

\bibitem{yuan_large-scale_2021}
Z.~Yuan, Y.~Yan, M.~Sonka, and T.~Yang, ``Large-scale robust deep auc maximization: {A} new surrogate loss and empirical studies on medical image classification,'' in \emph{Proceedings of the {IEEE}/{CVF} {International} {Conference} on {Computer} {Vision}}, 2021, pp. 3040--3049.

\bibitem{chong_image_2023}
C.~F. Chong, Y.~Wang, B.~Ng, W.~Luo, and X.~Yang, ``Image projective transformation rectification with synthetic data for smartphone-captured chest {X}-ray photos classification,'' \emph{Computers in Biology and Medicine}, p. 107277, 2023, publisher: Elsevier.

\bibitem{yuan_positive_2023}
Z.~Yuan, K.~Zhang, and T.~Huang, ``Positive {Label} {Is} {All} {You} {Need} for {Multi}-{Label} {Classification},'' \emph{arXiv preprint arXiv:2306.16016}, 2023.

\bibitem{kuznetsova_open_2020}
A.~Kuznetsova, H.~Rom, N.~Alldrin, J.~Uijlings, I.~Krasin, J.~Pont-Tuset, S.~Kamali, S.~Popov, M.~Malloci, A.~Kolesnikov, and {others}, ``The open images dataset v4: {Unified} image classification, object detection, and visual relationship detection at scale,'' \emph{International Journal of Computer Vision}, vol. 128, no.~7, pp. 1956--1981, 2020, publisher: Springer.

\bibitem{wang_saliency_2023}
S.~Wang, Q.~Wan, X.~Xiang, and Z.~Zeng, ``Saliency {Regularization} for {Self}-{Training} with {Partial} {Annotations},'' in \emph{Proceedings of the {IEEE}/{CVF} {International} {Conference} on {Computer} {Vision}}, 2023, pp. 1611--1620.

\bibitem{he_deep_2016}
K.~He, X.~Zhang, S.~Ren, and J.~Sun, ``Deep residual learning for image recognition,'' in \emph{Proceedings of the {IEEE} conference on computer vision and pattern recognition}, 2016, pp. 770--778.

\bibitem{tan_efficientnetv2_2021}
M.~Tan and Q.~Le, ``Efficientnetv2: {Smaller} models and faster training,'' in \emph{International {Conference} on {Machine} {Learning}}.\hskip 1em plus 0.5em minus 0.4em\relax PMLR, 2021, pp. 10\,096--10\,106.

\bibitem{woo_convnext_2023}
S.~Woo, S.~Debnath, R.~Hu, X.~Chen, Z.~Liu, I.~S. Kweon, and S.~Xie, ``Convnext v2: {Co}-designing and scaling convnets with masked autoencoders,'' in \emph{Proceedings of the {IEEE}/{CVF} {Conference} on {Computer} {Vision} and {Pattern} {Recognition}}, 2023, pp. 16\,133--16\,142.

\bibitem{dosovitskiy_image_2020}
A.~Dosovitskiy, L.~Beyer, A.~Kolesnikov, D.~Weissenborn, X.~Zhai, T.~Unterthiner, M.~Dehghani, M.~Minderer, G.~Heigold, S.~Gelly, and {others}, ``An image is worth 16x16 words: {Transformers} for image recognition at scale,'' \emph{arXiv preprint arXiv:2010.11929}, 2020.

\bibitem{liu_swin_2022}
Z.~Liu, H.~Hu, Y.~Lin, Z.~Yao, Z.~Xie, Y.~Wei, J.~Ning, Y.~Cao, Z.~Zhang, L.~Dong, and {others}, ``Swin transformer v2: {Scaling} up capacity and resolution,'' in \emph{Proceedings of the {IEEE}/{CVF} conference on computer vision and pattern recognition}, 2022, pp. 12\,009--12\,019.

\bibitem{chong_category-wise_2023}
C.~F. Chong, X.~Yang, T.~Wang, W.~Ke, and Y.~Wang, ``Category-{Wise} {Fine}-{Tuning} for {Image} {Multi}-label {Classification} with {Partial} {Labels},'' in \emph{International {Conference} on {Neural} {Information} {Processing}}.\hskip 1em plus 0.5em minus 0.4em\relax Springer, 2023, pp. 332--345.

\bibitem{cabral_matrix_2011}
R.~Cabral, F.~Torre, J.~P. Costeira, and A.~Bernardino, ``Matrix completion for multi-label image classification,'' \emph{Advances in neural information processing systems}, vol.~24, 2011.

\bibitem{xu_speedup_2013}
M.~Xu, R.~Jin, and Z.-H. Zhou, ``Speedup matrix completion with side information: {Application} to multi-label learning,'' \emph{Advances in neural information processing systems}, vol.~26, 2013.

\bibitem{yu_large-scale_2014}
H.-F. Yu, P.~Jain, P.~Kar, and I.~Dhillon, ``Large-scale multi-label learning with missing labels,'' in \emph{International conference on machine learning}.\hskip 1em plus 0.5em minus 0.4em\relax PMLR, 2014, pp. 593--601.

\bibitem{yang_improving_2016}
H.~Yang, J.~T. Zhou, and J.~Cai, ``Improving multi-label learning with missing labels by structured semantic correlations,'' in \emph{Computer {Vision}–{ECCV} 2016: 14th {European} {Conference}, {Amsterdam}, {The} {Netherlands}, {October} 11–14, 2016, {Proceedings}, {Part} {I} 14}.\hskip 1em plus 0.5em minus 0.4em\relax Springer, 2016, pp. 835--851.

\bibitem{wu_ml-mg_2015}
B.~Wu, S.~Lyu, and B.~Ghanem, ``Ml-mg: {Multi}-label learning with missing labels using a mixed graph,'' in \emph{Proceedings of the {IEEE} international conference on computer vision}, 2015, pp. 4157--4165.

\bibitem{kapoor_multilabel_2012}
A.~Kapoor, R.~Viswanathan, and P.~Jain, ``Multilabel classification using bayesian compressed sensing,'' \emph{Advances in neural information processing systems}, vol.~25, 2012.

\bibitem{gong_deep_2013}
Y.~Gong, Y.~Jia, T.~Leung, A.~Toshev, and S.~Ioffe, ``Deep convolutional ranking for multilabel image annotation,'' \emph{arXiv preprint arXiv:1312.4894}, 2013.

\bibitem{zhang_simple_2021}
Y.~Zhang, Y.~Cheng, X.~Huang, F.~Wen, R.~Feng, Y.~Li, and Y.~Guo, ``Simple and robust loss design for multi-label learning with missing labels,'' \emph{arXiv preprint arXiv:2112.07368}, 2021.

\bibitem{kim_bridging_2023}
Y.~Kim, J.~M. Kim, J.~Jeong, C.~Schmid, Z.~Akata, and J.~Lee, ``Bridging the {Gap} between {Model} {Explanations} in {Partially} {Annotated} {Multi}-label {Classification},'' in \emph{Proceedings of the {IEEE}/{CVF} {Conference} on {Computer} {Vision} and {Pattern} {Recognition}}, 2023, pp. 3408--3417.

\bibitem{bengio_curriculum_2009}
Y.~Bengio, J.~Louradour, R.~Collobert, and J.~Weston, ``Curriculum learning,'' in \emph{Proceedings of the 26th annual international conference on machine learning}, 2009, pp. 41--48.

\bibitem{chen_heterogeneous_2022}
T.~Chen, T.~Pu, L.~Liu, Y.~Shi, Z.~Yang, and L.~Lin, ``Heterogeneous {Semantic} {Transfer} for {Multi}-label {Recognition} with {Partial} {Labels},'' \emph{arXiv preprint arXiv:2205.11131}, 2022.

\bibitem{pu_semantic-aware_2022}
T.~Pu, T.~Chen, H.~Wu, and L.~Lin, ``Semantic-aware representation blending for multi-label image recognition with partial labels,'' in \emph{Proceedings of the {AAAI} conference on artificial intelligence}, vol.~36, 2022, pp. 2091--2098, issue: 2.

\bibitem{pu_dual-perspective_2023}
T.~Pu, T.~Chen, H.~Wu, Y.~Shi, Z.~Yang, and L.~Lin, ``Dual-{Perspective} {Semantic}-{Aware} {Representation} {Blending} for {Multi}-{Label} {Image} {Recognition} with {Partial} {Labels},'' 2023, \_eprint: 2205.13092.

\bibitem{yan_optimizing_2003}
L.~Yan, R.~H. Dodier, M.~Mozer, and R.~H. Wolniewicz, ``Optimizing classifier performance via an approximation to the {Wilcoxon}-{Mann}-{Whitney} statistic,'' in \emph{Proceedings of the 20th international conference on machine learning (icml-03)}, 2003, pp. 848--855.

\bibitem{cortes_auc_2003}
C.~Cortes and M.~Mohri, ``{AUC} optimization vs. error rate minimization,'' \emph{Advances in neural information processing systems}, vol.~16, 2003.

\bibitem{qi_stochastic_2021}
Q.~Qi, Y.~Luo, Z.~Xu, S.~Ji, and T.~Yang, ``Stochastic optimization of areas under precision-recall curves with provable convergence,'' \emph{Advances in Neural Information Processing Systems}, vol.~34, pp. 1752--1765, 2021.

\bibitem{montana_training_1989}
D.~J. Montana, L.~Davis, and {others}, ``Training feedforward neural networks using genetic algorithms.'' in \emph{{IJCAI}}, vol.~89, 1989, pp. 762--767.

\bibitem{gupta_comparing_1999}
J.~N. Gupta and R.~S. Sexton, ``Comparing backpropagation with a genetic algorithm for neural network training,'' \emph{Omega}, vol.~27, no.~6, pp. 679--684, 1999, publisher: Elsevier.

\bibitem{david_genetic_2014}
O.~E. David and I.~Greental, ``Genetic algorithms for evolving deep neural networks,'' in \emph{Proceedings of the {Companion} {Publication} of the 2014 {Annual} {Conference} on {Genetic} and {Evolutionary} {Computation}}, 2014, pp. 1451--1452.

\bibitem{pham_interpreting_2021}
H.~H. Pham, T.~T. Le, D.~Q. Tran, D.~T. Ngo, and H.~Q. Nguyen, ``Interpreting chest {X}-rays via {CNNs} that exploit hierarchical disease dependencies and uncertainty labels,'' \emph{Neurocomputing}, vol. 437, pp. 186--194, 2021, publisher: Elsevier.

\bibitem{huang_densely_2017}
G.~Huang, Z.~Liu, L.~Van Der~Maaten, and K.~Q. Weinberger, ``Densely connected convolutional networks,'' in \emph{Proceedings of the {IEEE} conference on computer vision and pattern recognition}, 2017, pp. 4700--4708.

\bibitem{deng_imagenet_2009}
J.~Deng, W.~Dong, R.~Socher, L.-J. Li, K.~Li, and L.~Fei-Fei, ``Imagenet: {A} large-scale hierarchical image database,'' in \emph{2009 {IEEE} conference on computer vision and pattern recognition}.\hskip 1em plus 0.5em minus 0.4em\relax Ieee, 2009, pp. 248--255.

\bibitem{chong_gan-based_2021}
C.~F. Chong, X.~Yang, W.~Ke, and Y.~Wang, ``{GAN}-based {Spatial} {Transformation} {Adversarial} {Method} for {Disease} {Classification} on {CXR} {Photographs} by {Smartphones},'' in \emph{2021 {Digital} {Image} {Computing}: {Techniques} and {Applications} ({DICTA})}.\hskip 1em plus 0.5em minus 0.4em\relax IEEE, 2021, pp. 01--08.

\bibitem{kingma_adam_2014}
D.~P. Kingma and J.~Ba, ``Adam: {A} method for stochastic optimization,'' \emph{arXiv preprint arXiv:1412.6980}, 2014.

\bibitem{gad_pygad_2021}
A.~F. Gad, ``{PyGAD}: {An} {Intuitive} {Genetic} {Algorithm} {Python} {Library},'' 2021, \_eprint: 2106.06158.

\bibitem{guo_fast_2020}
Z.~Guo, Y.~Yan, Z.~Yuan, and T.~Yang, ``Fast objective \& duality gap convergence for nonconvex-strongly-concave min-max problems,'' \emph{arXiv preprint arXiv:2006.06889}, 2020.

\bibitem{jansson_multi-view_2021}
P.~Jansson and {others}, ``Multi-view automated chest radiography interpretation,'' 2021.

\bibitem{kamal_anatomy_2021}
U.~Kamal, M.~Zunaed, N.~B. Nizam, and T.~Hasan, ``Anatomy {X}-{Net}: {A} {Semi}-{Supervised} {Anatomy} {Aware} {Convolutional} {Neural} {Network} for {Thoracic} {Disease} {Classification},'' \emph{arXiv preprint arXiv:2106.05915}, 2021.

\bibitem{ye_weakly_2020}
W.~Ye, J.~Yao, H.~Xue, and Y.~Li, ``Weakly supervised lesion localization with probabilistic-cam pooling,'' \emph{arXiv preprint arXiv:2005.14480}, 2020.

\bibitem{simonyan_very_2014}
K.~Simonyan and A.~Zisserman, ``Very deep convolutional networks for large-scale image recognition,'' \emph{arXiv preprint arXiv:1409.1556}, 2014.

\bibitem{cubuk_randaugment_2020}
E.~D. Cubuk, B.~Zoph, J.~Shlens, and Q.~V. Le, ``Randaugment: {Practical} automated data augmentation with a reduced search space,'' in \emph{Proceedings of the {IEEE}/{CVF} conference on computer vision and pattern recognition workshops}, 2020, pp. 702--703.

\bibitem{smith_super-convergence_2019}
L.~N. Smith and N.~Topin, ``Super-convergence: {Very} fast training of neural networks using large learning rates,'' in \emph{Artificial intelligence and machine learning for multi-domain operations applications}, vol. 11006.\hskip 1em plus 0.5em minus 0.4em\relax SPIE, 2019, pp. 369--386.

\bibitem{chen_multi-label_2019}
Z.-M. Chen, X.-S. Wei, P.~Wang, and Y.~Guo, ``Multi-label image recognition with graph convolutional networks,'' in \emph{Proceedings of the {IEEE}/{CVF} conference on computer vision and pattern recognition}, 2019, pp. 5177--5186.

\bibitem{chen_knowledge-guided_2020}
T.~Chen, L.~Lin, R.~Chen, X.~Hui, and H.~Wu, ``Knowledge-guided multi-label few-shot learning for general image recognition,'' \emph{IEEE Transactions on Pattern Analysis and Machine Intelligence}, vol.~44, no.~3, pp. 1371--1384, 2020, publisher: IEEE.

\bibitem{chen_learning_2021}
Z.~Chen, X.-S. Wei, P.~Wang, and Y.~Guo, ``Learning graph convolutional networks for multi-label recognition and applications,'' \emph{IEEE Transactions on Pattern Analysis and Machine Intelligence}, 2021, publisher: IEEE.

\bibitem{lin_focal_2017}
T.-Y. Lin, P.~Goyal, R.~Girshick, K.~He, and P.~Dollár, ``Focal loss for dense object detection,'' in \emph{Proceedings of the {IEEE} international conference on computer vision}, 2017, pp. 2980--2988.

\end{thebibliography}
\bibliographystyle{IEEEtran}


 





\end{document}